\documentclass[lettersize,journal]{IEEEtran}
\usepackage{amsmath,amsfonts}
\usepackage{array}
\usepackage[caption=false,font=normalsize,labelfont=sf,textfont=sf]{subfig}
\usepackage{textcomp}
\usepackage{stfloats}
\usepackage{url}
\usepackage{verbatim}
\usepackage{graphicx}
\usepackage{cite}
\usepackage{amssymb}
\usepackage{amsmath}
\usepackage{amsfonts}
\usepackage{makeidx}
\usepackage{graphicx}
\usepackage{epstopdf}
\usepackage{diagbox}
\usepackage{booktabs}
\usepackage[usenames,dvipsnames]{color}
\usepackage{multirow}
\usepackage{url}

\usepackage[misc]{ifsym}

\usepackage{multirow}%
\usepackage{amsmath,amssymb,amsfonts}%
\usepackage{amsthm}%
\usepackage{mathrsfs}%
\usepackage{xcolor}%
\usepackage{textcomp}%
\usepackage{booktabs}
\usepackage{graphicx}
\usepackage{setspace}
\usepackage{manyfoot}%
\usepackage{algorithmicx}%
\usepackage[lined,boxed,commentsnumbered]{algorithm2e}
\usepackage{algpseudocode}%
\usepackage{listings}%

\usepackage{array}

\newcommand{\gensimp}[1]{\sigma_{#1}}

\newcommand{\genopea}[2]{\boldsymbol{\partial}^{#2}_{#1}}
\newcommand{\genasgn}[1]{\mathbf{B}_{#1}}

\newcommand{\genhlop}[2]{\boldsymbol{\mathcal{L}}^{#2}_{#1}}
\newcommand{\genbase}[2]{\psi_{#1}^{#2}}
\newcommand{\geneign}[2]{\lambda_{#1}^{#2}}
\newcommand{\gencoef}[2]{c_{#1}^{#2}}

\def\mygraph{\mathcal{G}}
\def\myfilter{\boldsymbol{h}}
\def\mysignal{\boldsymbol{x}}

\newcommand{\mymatrix}[1]{\boldsymbol{X}_{#1}}
\newcommand{\myweight}[2]{\boldsymbol{W}^{#2}_{#1}}

\newcommand{\mymatprm}[1]{\tilde{\boldsymbol{X}}_{#1}}
\newcommand{\mymatatt}[1]{\boldsymbol{A}_{#1}}

\newcommand{\genspec}[1]{\boldsymbol{\lambda}_k}
\newcommand{\genpoly}[1]{T_{#1}}
\newcommand{\genexpa}[1]{\theta_{#1}}

\newcommand{\opprjdw}[1]{\mathbf{T}_{ \mathrm{#1\downarrow} }}
\newcommand{\opprjup}[1]{\mathbf{T}_{ \mathrm{\uparrow #1} }}
\newcommand{\opsstdw}[2]{\mathbf{T}_\mathrm{#1\downarrow #2}}
\newcommand{\opsstup}[2]{\mathbf{T}_\mathrm{#1 \uparrow #2}}

\def\ka{k_1}
\def\kb{k_2}
\newcommand{\opbound}[1]{\mathbf{J}_{\cdot,#1}}

\begin{document}

\title{Advancing Graph Neural Networks with HL-HGAT: A Hodge-Laplacian and Attention Mechanism Approach for Heterogeneous Graph-Structured Data}

\author{Jinghan Huang*,
        Qiufeng Chen*,
        Yijun Bian,
        Pengli Zhu,~\IEEEmembership{Student Member,~IEEE,}
        Nanguang Chen,
        Moo K. Chung,
        Anqi Qiu,~\IEEEmembership{Member,~IEEE}
\thanks{Jinghan Huang, Yijun Bian, Pengli Zhu, Nanguang Chen, and Anqi Qiu are with the Department of Biomedical Engineering, National University of Singapore (e-mails: e0917898@u.nus.edu; yjbian@nus.edu.sg; dlmu.p.l.zhu@gmail.com; biecng@nus.edu.sg).}
\thanks{Anqi Qiu is also with the Department of Health Technology and Informatics, the Hong Kong Polytechnic University (e-mail: an-qi.qiu@polyu.edu.hk).}
\thanks{Qiufeng Chen is with the College of Computer and Information Science, 
Fujian Agriculture and Forestry University, Fuzhou, China (e-mail: chenqiufeng0204@126.com).}
\thanks{Moo K. Chung is with the Department of Biostatistics and Medical Informatics, the University of
Wisconsin-Madison, Wisconsin, USA (e-mail: mkchung@wisc.edu).}
\thanks{*These authors contributed equally to this work.}
\thanks{Correspondence to Anqi Qiu (an-qi.qiu@polyu.edu.hk).}
}

\maketitle

\begin{abstract}
Graph neural networks (GNNs) have proven effective in capturing relationships among nodes in a graph. This study introduces a novel perspective by considering a graph as a simplicial complex, encompassing nodes, edges, triangles, and $k$-simplices, enabling the definition of graph-structured data on any $k$-simplices. Our contribution is the Hodge-Laplacian heterogeneous graph attention network (HL-HGAT), designed to learn heterogeneous signal representations across $k$-simplices. The HL-HGAT incorporates three key components: HL convolutional filters (HL-filters), simplicial projection (SP), and simplicial attention pooling (SAP) operators, applied to $k$-simplices. HL-filters leverage the unique topology of $k$-simplices encoded by the Hodge-Laplacian (HL) operator, operating within the spectral domain of the $k$-th HL operator. To address computation challenges, we introduce a polynomial approximation for HL-filters, exhibiting spatial localization properties. Additionally, we propose a pooling operator to coarsen $k$-simplices, combining features through simplicial attention mechanisms of self-attention and cross-attention via transformers and SP operators, capturing topological interconnections across multiple dimensions of simplices. The HL-HGAT is comprehensively evaluated across diverse graph applications, including NP-hard problems, graph multi-label and classification challenges, and graph regression tasks in logistics, computer vision, biology, chemistry, and neuroscience. The results demonstrate the model's efficacy and versatility in handling a wide range of graph-based scenarios.

\end{abstract}


\section{Introduction}
\label{sec:introduction}

\IEEEPARstart{G}{raph}-structured data, embodying entities as nodes and their relationships as edges, offer a robust framework for modeling and analyzing complex interdependencies. To harness the potent representations of these data, graph neural networks (GNNs) have been engineered, primarily focusing on node-centric message aggregation to discern features defined on graph nodes \cite{zhou2020graph,wu2020comprehensive}. Such an approach has proved valuable across multiple domains, illuminating insights and driving informed decisions \cite{kipf2016semi,velivckovic2017graph}. 
The versatility and utility of GNNs stand out, whether it be in predicting molecular generation and chemical properties \cite{reiser2022graph,gilmer2017neural},  analyzing brain networks \cite{huang2022spatio,cui2022braingb}, forecasting traffic flow in transport systems  \cite{peng2020spatial,jiang2022graph}, or numerous other applications \cite{huang2021knowledge, wu2022graph,hanocka2019meshcnn,alonso2021graph,pradhyumna2021graph}. 
However, the real world frequently exhibits heterogeneous data defined on nodes, edges, and more intricate connections among these graph elements \cite{peng2005technologies,farazi2023tetcnn}. Such scenarios reveal an extant gap: the need for a generic GNN capable of comprehending and analyzing higher-order relationships and dependencies within heterogeneous graph-structured data. 

The architecture of existing GNNs, analogous to conventional convolutional neural networks (CNNs) \cite{NIPS1989_53c3bce6}, encompasses two principal components: convolution over graph nodes and graph pooling for spatial dimensionality reduction and multi-scale learning \cite{wu2020comprehensive}. However, the irregular topology of graphs poses challenges to the shifting operation inherent to convolution on a graph. Current GNNs attempt to navigate this in two ways, through the spatial and spectral domains.  In the spatial domain, GNNs endeavor to unearth valuable information from node neighborhoods to facilitate message aggregation for node convolution.  Graph convolutional network (GCN) \cite{kipf2016semi} and dynamic graph convolutional network (dGCN) \cite{zhao2022dynamic} initiate graph convolution with isotropic normalization filters, but they lack adaptive node weighting.
Some elaborate mechanisms are introduced into convolutional layers to allow for adaptive aggregation of neighborhood information into target nodes \cite{velivckovic2017graph, li2021braingnn, bresson2017residual, rampavsek2022recipe}, such as clustering-based embedding mechanisms in BrainGNN \cite{li2021braingnn}, attention mechanisms in graph attention network (GAT) \cite{velivckovic2017graph},  general, powerful, and scalable (GPS) graph transformer \cite{rampavsek2022recipe}, and gating mechanisms in GatedGCN (gated graph convolutional network) \cite{bresson2017residual}. 
Especially, the attention mechanism via a transformer \cite{vaswani2017attention} perceives each node as a token and learning weights over nodes of a graph for node-centric message aggregation \cite{velivckovic2017graph,rampavsek2022recipe}.

In the spectral domain, graph convolution is achieved through the design of spectral filters via the graph Laplacian, where filter values represent the importance of neighborhood nodes \cite{bruna2014spectral}. For computational efficiency, Chebyshev polynomials and other polynomial forms are used to approximate spectral filters for GNNs \cite{defferrard2016convolutional, huang2021revisiting}. When dealing with larger graphs, spectral graph convolution with polynomial approximation emerges as a computationally efficient and spatially localized approach \cite{huang2021revisiting}.

Despite their widespread use, node-centric convolutions are not adept at managing high-order dimensional relationships between nodes,  edges, and beyond. Thus far,  CensNet \cite{jiang2019censnet} and Hypergraph NN \cite{jo2021edge} have employed techniques,  such as line graph transformation and dual hypergraph transformation,  to interchange the roles of nodes and edges within a graph. Spatial graph convolution on simplicial complexes, such as nodes and edges, is also introduced via Hodge-Laplace operators without fast numerical solutions \cite{courtney2016generalized,barbarossa2020topological,bodnar2021weisfeiler}. Giusti et al. further incorporated a self-attention mechanism into the convolution operator \cite{giusti2022simplicial}. This allows for the application of traditional GNN convolution beyond nodes.  These models have demonstrated significant success in social networks,  molecular science, and neuroscience \cite{monti2018dual, jiang2019censnet, jo2021edge, huang2023heterogeneous}, suggesting that edge-based convolutions can be advantageous when signals are inherently associated with edges.



The second crucial component of GNNs is graph pooling, which serves to reduce data dimensionality and learn multi-scale spatial information \cite{grattarola2022understanding}. Currently, most graph pooling methods first coarsen a graph by clustering nodes and then pool signals by taking their average or maximum across the nodes within each cluster \cite{defferrard2016convolutional, ying2018hierarchical}. An alternative approach is to perform graph pooling by discarding nodes based on their attention scores \cite{gao2019graph, gao2021topology}.  Cinque et al. \cite{cinque2023pooling} further extended pooling methods from nodes to edges.
However, to the best of our knowledge, no existing graph pooling methods consider higher-order interactions between nodes,  edges,  or beyond during signal pooling.  As a result, there is a pressing need for innovative graph pooling methods that are efficient, effective, and capable of leveraging heterogeneous information on nodes, edges, and beyond.

In this study,  we present a pioneering approach, named the Hodge-Laplacian heterogeneous graph attention network (HL-HGAT). This model interprets a graph as a simplicial complex,  encompassing nodes, edges, triangles, and more,  enabling the definition of graph-structured data on any $k$-simplex.  In this setting,  the HL-HGAT aims to learn heterogeneous signal representations using Hodge-Laplacian (HL) operators on  $k$-simplices while capturing their intricate complex relationships across $k$-simplices.  We propose the HL-HGAT with three key elements: convolutional filters constructed by HL operators,  multi-simplicial interaction (MSI) by introducing simplicial projection operators,  and simplicial attention pooling (SAP) via simplicial transformers. 
Notably, we utilize the $k$-th boundary operator encoding the relationship between the $(k-1)$-simplices and $k$-simplices,  enabling the construction of the $k$-th HL operator on $k$-simplices.  We then develop convolutional filters on the $k$-simplices in the spectral domain of the $k$-th HL operator,  termed HL-filters. This necessitates the computation of the $k$-th HL eigenfunctions,  which can be computationally demanding for larger graphs.  To mitigate this, we introduce a generic polynomial approximation of the HL-filters, which exhibits a spatial localization property relative to the polynomial order, representing a significant advancement in the field.  Furthermore, we design a simplicial projection operator that facilitates the conversion of signals from $\ka$-simplices to $\kb$-simplices, enabling their fusion and thereby enabling the learning of their interactions.  In addition, we define a pooling operator by coarsening the $k$-simplices and consolidating features associated with these simplices using attention mechanisms. These mechanisms encompass self-attention and cross-attention through the simplicial projection operators and transformers, collectively referred to as simplicial transformers. The simplicial transformers ascertain the importance of each simplex by learning its weight while gathering signals from its topologically connected simplices, thereby assessing their relevance to a downstream task.  Therefore,  our innovative solution---HL-HGAT---encompasses the development of HL-filters, simplicial projection operators, and simplicial transformers, providing a comprehensive framework for capturing complex relationships in graph-structured data.


To rigorously assess the HL-HGAT model, we conducted comprehensive evaluations across a diverse range of graph applications, including NP-hard, graph multi-label and classification, and graph regression problems in the domains of logistics, computer vision, biology, chemistry, and neuroscience. 
We compared the HL-HGAT performance with state-of-the-art GNN models, such as GCN \cite{kipf2016semi}, GAT \cite{velivckovic2017graph},  GatedGCN \cite{bresson2017residual},  and GPS \cite{rampavsek2022recipe}, as well as GNN models specialized for brain network data,  such as BrainGNN \cite{li2021braingnn},  dGCN \cite{zhao2022dynamic}, Hypergraph NN \cite{jo2021edge},  using six benchmark datasets.  Our results include interpretable attention maps that demonstrate the HL-HGAT's ability to learn meaningful representations at nodes,  edges, and beyond.

The rest of this paper is organized as follows. 
We summarize the related work in Section~\ref{sec:related}, detail each component of the proposed HL-HGAT in Section~\ref{sec:method}, and evaluate the performance of HL-HGAT in Section~\ref{sec:empirical}, respectively.


\section{Related Work}
\label{sec:related}

\subsection{Convolution on a Graph}

\noindent{\bf{Spatial Domain.}} Graph convolution in the spatial domain is a prominent technique in graph neural networks, providing a mechanism to map signals on nodes into latent features by leveraging information from their respective neighborhoods.  The concept of graph convolution was first introduced using isotropic normalization filters to aggregate neighborhood signals with equal weight, as illustrated by early works such as 
graph convolutional network (GCN) \cite{kipf2016semi} and dynamic graph convolutional network (dGCN) \cite{zhao2022dynamic}. 
These methods were relatively straightforward and relied on the premise that every node in the neighborhood contributes equally to the feature representation of the target node. Although it provided a starting point for graph convolution, it lacked the capability to capture the varying importance of different nodes.

Recognizing the limitation of equal node contribution, subsequent development introduced various weight updating strategies, accounting for signal differences between nodes. These strategies incorporated attention or gating mechanisms to discern the differences between nodes and assign weights accordingly \cite{velivckovic2017graph,bresson2017residual}. The inclusion of these mechanisms allowed for the realization of anisotropic convolutions, leading to more nuanced and effective aggregation of neighborhood information.  For instance, the gated graph convolutional network (GatedGCN) developed by Bresson and Laurent (2017) leverages a gating mechanism to introduce anisotropy into the convolution process \cite{bresson2017residual}. The gating mechanism, inspired by recurrent neural networks (RNNs), facilitates dynamic weight assignment based on the relationships between nodes, ensuring a more accurate and adaptive aggregation process. Similarly, the graph attention network (GAT) utilizes a transformer architecture to learn attention coefficients that assign different weights to neighboring nodes based on their importance \cite{velivckovic2017graph}. The attention mechanism allows the model to focus on relevant nodes, leading to more efficient feature extraction.

In addition to the previously mentioned techniques, there are other innovative approaches such as the Graph Transformer, which treats each node as a token and encodes its position along with signals on nodes  \cite{vaswani2017attention}. This method allows for the capture of global dependencies between nodes, enhancing the expressive power of graph convolution. A more recent approach is the general, powerful, and scalable (GPS) graph transformer \cite{rampavsek2022recipe}. This model blends the strengths of traditional transformers with spatial graph convolution in each layer to effectively learn signal representation over nodes. This innovative approach encapsulates the benefits of spatial-based graph convolution techniques, providing a robust tool for handling graph-structured data defined on nodes of graphs.

\smallskip
\noindent{\bf{Spectral Domain.}} In the spectral domain, graph convolution can be achieved via the graph Laplacian that captures the graph's topological characteristics \cite{bruna2014spectral}. The groundbreaking work of Defferrard et al. revolutionized the application of convolution operations to graphs in the spectral domain \cite{defferrard2016convolutional}. They developed a fast localized convolutional filter that propagates information within the $k$-hop neighborhood of a node.  Here, a ``hop'' signifies the adjacency relationship between nodes: two nodes are within one hop of each other if they are directly connected by an edge, within two hops if they are connected via one intermediary node, and so forth. The $k$-hop neighborhood of a node encompasses all nodes reachable within $k$ hops.  To enhance the efficiency of this convolution operation,  a parameterization based on Chebyshev polynomials is proposed \cite{defferrard2016convolutional,huang2021revisiting}. Polynomials (e.g., Chebyshev, Hermite, Laguerre) are a sequence of orthogonal polynomials, which are well-suited to approximation tasks \cite{huang2021revisiting}. By expressing the graph convolution operation in terms of polynomials,  the graph convolution achieves linear computation complexity. This development was a significant contribution, as it made the convolution operation computationally feasible for large-scale graphs.

However, the state-of-the-art in graph convolution is not without limitations. The convolution operations focus primarily on nodes and the messages are passed solely among these entities. This methodology overlooks the rich information contained in higher-dimensional simplices, such as edges (1-simplices) and triangles (2-simplices). Edges, for instance, embody crucial relationships between nodes, and their information may be valuable in many graph-related tasks. Hence, future research in the spectral domain of graph convolution could benefit significantly from exploring methods that can effectively utilize information from these higher-dimensional simplices, which is the focus of the present study.

\subsection{Graph Pooling}
Graph pooling is a crucial operation in GNNs that mirrors the function of pooling layers in traditional CNNs \cite{NIPS1989_53c3bce6}. Its primary objective is to reduce the spatial dimensionality of graph data while preserving the essential structural information. One common approach to graph pooling is to group nodes into clusters and then aggregate, or pool, the signals within each cluster. This aggregation can be achieved by taking the average or the maximum signal value within each cluster, which effectively reduces the dimensionality while capturing the intra-cluster relationships  \cite{defferrard2016convolutional}.  Similarly, MinCutPool, inspired by spectral clustering, encourages nodes within the same cluster to be strongly connected and have similar features by utilizing a minCUT loss \cite{bianchi2020spectral}.

Differentiable graph pooling (DIFFPOOL) softens the assignment of nodes to clusters by using a learned assignment matrix that probabilistically maps nodes to clusters in a coarsened graph \cite{ying2018hierarchical}.   This process makes the pooling operation differentiable and facilitates end-to-end training of the GNN. However, this approach can be both memory and computationally expensive as it requires learning a dense assignment matrix. 

Another approach to graph pooling is to discard a fixed proportion of nodes based on some criterion. For example, TopKPool applies a linear layer to project node features into scalars and then ranks them, selecting the $k$ nodes with the largest values \cite{gao2019graph}. Similarly, topology-aware pooling (TAPool) generates scores from both local and global voting and selects nodes based on these scores \cite{gao2021topology}.

Despite the considerable progress in graph pooling, a notable gap remains in applying these techniques to $k$-simplices, which are generalizations of nodes (0-simplex), edges (1-simplex), triangles (2-simplex), and so on. These higher-order structures contain rich information about the topological properties of the graph, which is underutilized in current graph pooling methods. The challenge of effectively pooling such complex, multi-dimensional structures is our research to generalizing GNNs on $k$-simplices.

\section{Methods}
\label{sec:method}

In this section, we provide a comprehensive exploration of the mathematical foundation that underlies Hodge-Laplacian spectral filters (HL-filters) and simplicial attention pooling (SAP) operators when applied to $k$-simplices,  as well as multi-simplicial interaction (MSI) across multi-dimensional simplices.  Within the overarching framework of HL-HGAT, we shed light on the HL-filters' association with $k$-simplices through the utilization of the $k$-th HL operator, accompanied by an in-depth discussion of their polynomial approximations. Subsequently, we introduce the simplicial projection (SP) operator, designed to facilitate signal transformation between simplices of different dimensions, while also introducing the concept of MSI to enable the learning of signal representations spanning across multi-dimensional simplices. This discussion is further enriched with the introduction of simplicial attention pooling, a novel approach that leverages attention mechanisms to distill features, both intrinsically within a specific dimensional simplex and interdimensionally across a multitude of simplices.

\subsection{Spectral Filters via the Hodge-Laplacian Operator (HL-filters)} \label{HLfilter}

\begin{figure*}[htbp]
\begin{center}
\includegraphics[width=0.9\textwidth]{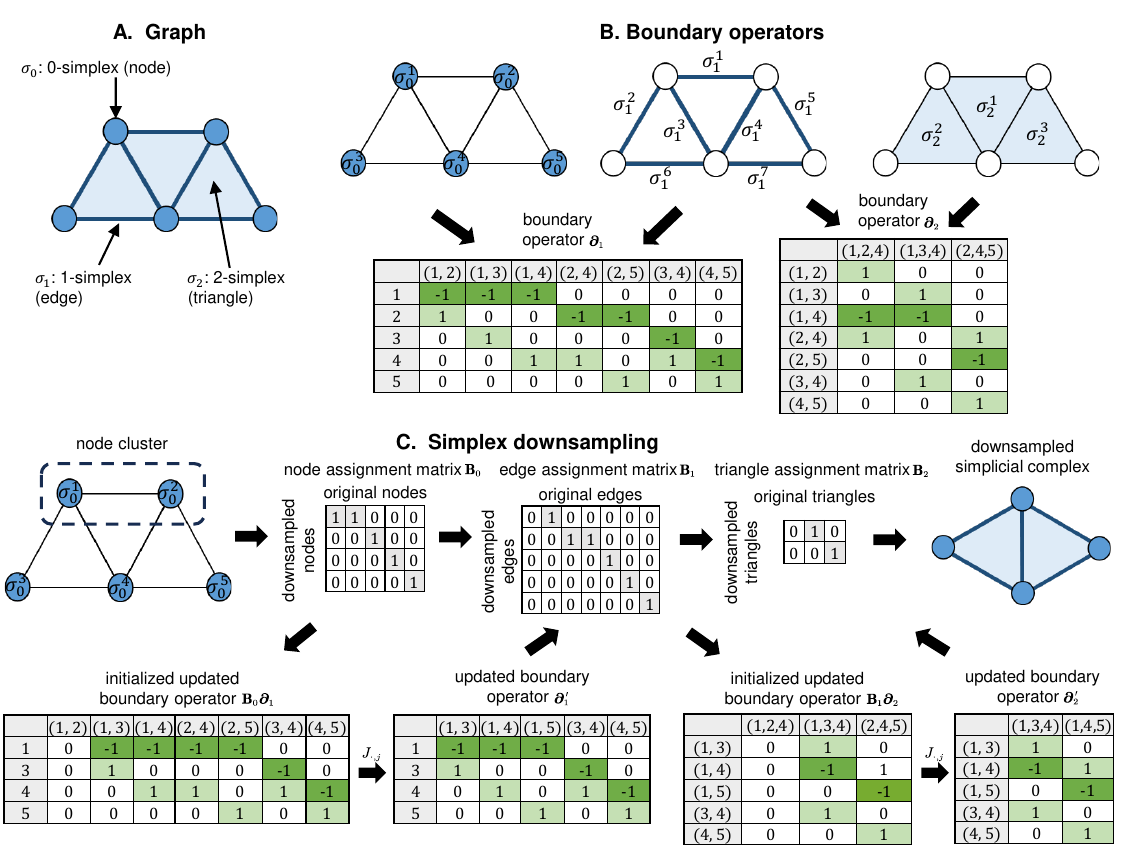}
\caption{Illustration of simplices,  boundary operators,  and simplex downsampling.
Panel (A) illustrates a graph with the $0$-, $1$-, $2$-simplices, while Panel (B) displays its corresponding $1$-st and $2$-nd boundary operators. 
Panel (C) demonstrates an example of the simplex downsampling and the update of the corresponding boundary operators. In the simplex assignment matrix, each row corresponds to the $k$-simplex of the downsampled graph, and each column corresponds to the $k$-simplex of the original graph. }  
\label{fig:simplex} 
\end{center}
\end{figure*}

Let's denote a graph by $\mygraph$ as a simplicial complex comprising nodes, edges, triangles, and $k$-simplices, as shown in Fig. \ref{fig:simplex}(A). 
A node on $\mygraph$ corresponds to a $0$-simplex, denoted as $\gensimp{0}$; An edge on $\mygraph$, connecting two nodes, represents a $1$-simplex, denoted as $\gensimp{1}$; A $k$-simplex $\gensimp{k}$ is defined as a $k$-th polytope with $(k+1)$ nodes. 
In this study,  we utilize the $k$-th  boundary operator $\genopea{k}{}$ to characterize the relationship between the $(k-1)$-simplex and $k$-simplex on $\mygraph$. 
For $k\in \mathbb{Z}^+$, this $k$-th boundary operator is expressed in a matrix form as 
\begin{equation}
\begin{bmatrix}\genopea{k}{}\end{bmatrix}_{ij}= 
\begin{cases}
    \hspace{.8em} 1\,, & \text{if} \; \gensimp{k-1}^i \; \text{is positively oriented w.r.t. } \gensimp{k}^j \,;\\
    -1\,,& \text{if} \; \gensimp{k-1}^i \; \text{is negatively oriented w.r.t. } \gensimp{k}^j \,;\\
    \hspace{.8em} 0\,, & \text{otherwise}\,,
\end{cases}%
\end{equation}%
The element of $\genopea{k}{}$ in the $i$-th row and $j$-th column 
indicates whether the $i$-th $(k-1)$-simplex $\gensimp{k-1}^i$ belongs to the $j$-th $k$-simplex $\gensimp{k}^j$. 
An illustrative example is given in Fig. \ref{fig:simplex}(B) to present the construction of the boundary operators.

\begin{figure*}[!htbp]
\begin{center}
\includegraphics[width=0.92\textwidth]{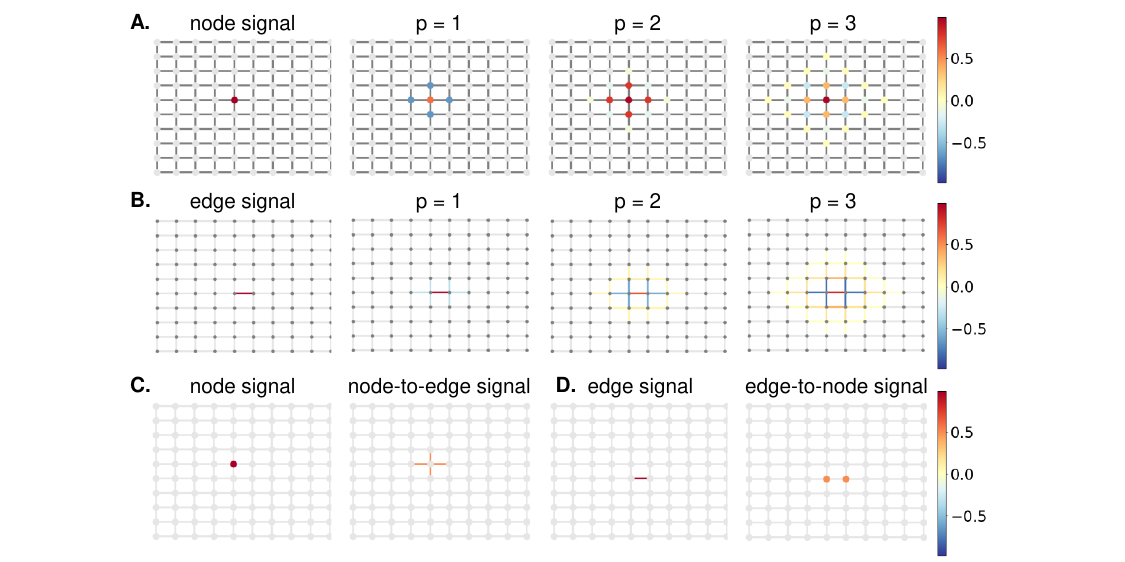}
\caption{HL-filters and simplicial projection operators. 
(A--B) The series of panels, from left to right, display a pulse signal associated with either a node or an edge, followed by the corresponding signals filtered through HL-filters employing approximations based on the 1st, 2nd, and 3rd-order Laguerre polynomials. 
(C--D) The left and right panels respectively depict a pulse signal defined on either a node or an edge and their resulting signals after projection through a simplicial projection operator.
 }  
\label{HL_filter_size} 
\end{center}
\end{figure*} 
Specifically, if we assume that a graph $\mathcal{G}$ has $n_0$ nodes, 
it is worth noting that the element $\begin{bmatrix}\genopea{1}{}\end{bmatrix}_{ij}$ encodes whether the $i$-th and $j$-th nodes (i.e., $0$-simplices) are connected to form an edge (i.e., a $1$-simplex) \cite{edelsbrunner2000topological}. 
In graph theory \cite{lee2014hole}, the $1$-st boundary operator $\genopea{1}{}$ is equivalent to a traditional incidence matrix with size of $n_0\times \frac{n_0\times (n_0-1)}{2}$, where nodes are indexed over rows and edges are indexed over columns. 
When $\begin{bmatrix}\genopea{1}{}\end{bmatrix}_{ij}$ is equal to $\pm 1$, the $i$-th and $j$-th nodes form an edge that connects these two nodes. Similarly, the $2$-nd boundary operator $\genopea{2}{}$ encodes how $1$-simplices (i.e., nodes) are connected to form a $2$-simplex (i.e., a triangle). 

We now design spectral filters via the $k$-th Hodge-Laplacian (HL) operator on the $k$-simplices.  The $k$-th HL operator is defined as 
\begin{equation}
\label{eqn:hlopt}
\genhlop{k}{}=
\genopea{k+1}{} \genopea{k+1}{\top}+
\genopea{k}{\top} \genopea{k}{} \,.
\end{equation}
Note that a special case of $k=0$ is the $0$-th HL operator over notes, that is,
\begin{equation}
\label{eqn:L0}
\genhlop{0}{}=
\genopea{1}{} \genopea{1}{\top} 
\,.
\end{equation}
This special case is equivalent to the standard graph Laplacian operator, i.e., $\genhlop{0}{} \equiv \Delta$. 
Then by solving the eigensystem of the $k$-th HL operator, we can obtain a set of orthonormal bases $\{\genbase{k}{0}, \genbase{k}{1}, \genbase{k}{2}, \cdots\}$ on the $k$-simplices, that is, 
\begin{equation}
    \genhlop{k}{} \genbase{k}{j} =
    \geneign{k}{j} \genbase{k}{j} \,,
\end{equation}%
where $\geneign{k}{j}$ represents the eigenvalue corresponding to the eigenvector $\genbase{k}{j}$, and $\genspec{k}= [\geneign{k}{0}, \geneign{k}{1}, \geneign{k}{2}, \cdots]^\top$ is the spectrum of the matrix $\genhlop{k}{}$. 

Now we consider an HL-filter $\myfilter(\cdot)$ with spectrum as
\begin{equation}
    \myfilter(\cdot,\cdot)=
    \sum_{j=0}^\infty
    \myfilter(\geneign{k}{j})
    \genbase{k}{j}(\cdot)
    \genbase{k}{j}(\cdot) 
    \,,\label{eq:hy}
\end{equation}%
then a generic form of convolution of a signal $\mysignal$ with a HL-filter $\myfilter(\cdot)$ on the heterogeneous graph $\mygraph$ can be defined by firstly transforming the signal to the spectral domain and then filtering it based on the corresponding spectrum of $\myfilter$ \cite{tan2014spectral,huang2021revisiting}, that is,
\begin{equation}
    \mysignal'(\cdot)= 
    \myfilter \ast \mysignal(\cdot)=
    \sum_{j=0}^\infty
    \myfilter(\geneign{k}{j})
    \gencoef{k}{j}
    \genbase{k}{j}(\cdot) 
    \,,\label{eq:hx}
\end{equation}%
where $\mysignal(\cdot)= \sum_{j=0}^\infty \gencoef{k}{j}\genbase{k}{j}(\cdot)$. 
Specifically, the signal $\mysignal$ is defined on the nodes of the graph $\mygraph$ when $k=0$. 

\subsection{Polynomial Approximation of HL-filters}

The shape of a HL-filter (e.g., $\myfilter(\cdot)$ in Eq.~\eqref{eq:hx}) 
determines how many simplices are aggregated in the filter process. When $\mygraph$ is large, the computation of $\geneign{k}{j}$ and $\genbase{k}{j}$ is intensive. 
To mitigate the costly computation, 
in this subsection, we propose to approximate a HL-filter by the expansion of polynomials, that is, $\{\genpoly{p}\mid p=0,1,2,\dots,P-1\}$, such that 

\begin{equation}
\label{eq:hlambda}
\myfilter(\genspec{k})= \sum_{p=0}^{P-1} 
\genexpa{p} \genpoly{p}(\genspec{k}) \,,
\end{equation}%
where $\genexpa{p}$ is the expansion coefficient associated with the $p^\text{th}$-order polynomial. The coefficients $\{\genexpa{p} \mid p=0,1,...,P-1\}$ are the parameters of the HL-filter for optimization. 

We can rewrite the convolution in Eq. (\ref{eq:hx}) as: 
\begin{equation}
\label{eq:gx3}
\mysignal'(\cdot)= \myfilter\ast \mysignal(\cdot)=
\sum_{p=0}^{P-1} \genexpa{p}
\genpoly{p}(\genhlop{k}{})
\mysignal(\cdot) \,,
\end{equation}%
where $\genpoly{p}$ can be any polynomial that is represented by a recursive equation. 
In this study, we select the Laguerre polynomial \cite{822801, huang2020fast}, which can be computed from a recurrence relation, that is, 
\begin{equation}
    \genpoly{p+1}(\genspec{k}) =\frac{
        (2p+1-\genspec{k}) \genpoly{p}(\genspec{k})
        -p\genpoly{p-1}(\genspec{k})
    }{p+1} \,,
\end{equation}%
with $\genpoly{0}(\genspec{k}) =1$ and $\genpoly{1}(\genspec{k}) =1-\genspec{k}$. 

Fig. \ref{HL_filter_size}(A, B) provide clear illustrations of filtered node and edge pulse signals, showcasing the results of approximating Laguerre polynomials of different orders, such as the 1st, 2nd, and 3rd-orders. This visualization effectively demonstrates that the spatial localization property of the HL-filters is determined by the order of Laguerre polynomials.
Notably, alternative polynomials with a recurrence relation can also be employed for approximating the HL-filter as shown in Huang et al. \cite{huang2021revisiting}. 

\subsection{Simplicial Projection Operator}
Signals on simplices of different dimensions may be relevant to each other. However, the HL-filters described above work exclusively on one specific-dimensional simplex, in other words, they do not operate across multiple dimensions. 
To address this problem, we define two projection operators, that is, $\opprjdw{k}$ and $\opprjup{k}$. 
The first projection operator $\opprjdw{k}$ is supposed to map signals from the $k$-simplices to the $(k-1)$-simplices. The design of this operator relies on the topological relationship between the $k$- and $(k-1)$-simplices of a graph $\mygraph$, and this relationship is captured by the $k$-th boundary operator of $\mygraph$. 
As a result, $\opprjdw{k}$ is expressed as 
\begin{equation}
   \begin{bmatrix} \opprjdw{k}\end{bmatrix}_{ij} = 
        \left|\begin{bmatrix}\genopea{k}{}\end{bmatrix}_{ij} \right|
    =\begin{cases}
        1 \,,&\text{if } \gensimp{k-1}^i \text{ is incident with } \gensimp{k}^j \,;\\
        0 \,,&\text{otherwise} \,,
    \end{cases}
    \label{Tk:b}
\end{equation}%
%
where the notation of $| \cdot |$ represents the absolute value function. Similarly, the second projection operator $\opprjup{k}$ is supposed to map signals from the $(k-1)$-simplices to the $k$-simplices, and it can be rewritten as the transpose of $\opprjdw{k}$, that is,
\begin{equation}
    \opprjup{k} = \opprjdw{k}^\top
    \,.\label{Tk:c}
\end{equation}%
The pair of these two projection operators will serve as the fundamental operators to facilitate the interaction of signals across multi-dimensional simplices. 

We now define a generic projection operator to map signals from the $\kb$-simplices to the $\ka$-simplices, where $\ka < \kb$. 
This mapping is performed iteratively by projecting signals: first from $\kb$ to $(\kb -1)$, then from $(\kb -1)$ to $(\kb -2)$, and so on. Then this iterative projection could be denoted by
\begin{equation}
    \opsstdw{\kb}{\ka}= 
    \opprjdw{\ka+1} \cdots \opprjdw{\kb-1} \opprjdw{\kb} 
    =\prod_{k=\ka+1}^{\kb} \opprjdw{k}
    \,.\label{eq:T12}
\end{equation}%
Conversely, to project signals from $\ka$-simplices to $\kb$-th ones, we use the operator
\begin{equation}
    \opsstup{\ka}{\kb}=
    \opprjup{\kb} \cdots \opprjup{\ka+2} \opprjup{\ka+1} 
    =\prod_{k=\kb}^{\ka+1} \opprjup{k}
    =\opsstdw{\kb}{\ka}^\top
    \,.\label{eq:T21}
\end{equation}%

Fig.~\ref{HL_filter_size}(C,D) provide a practical simulation of projecting a pulse signal defined on a node to its neighboring edges and a pulse signal defined on an edge to its neighboring nodes. The proposed simplicial projection operator can map any simplex's signal with any dimensional neighboring simplices' signals, which enables us to learn signal interactions and cross-attentions across multiple dimensional simplices introduced in Section~\ref{appx:msi}.

\subsection{Multi-simplicial Interaction (MSI)}
\label{appx:msi}
Based on the simplicial projection operator, we define a multi-simplicial interaction layer. This layer is tailored to learn signal representations on the $k$-simplices, incorporating influences from signals on other dimensional simplices. Assume we have a signal $\mysignal$ 
defined on the $k$-simplices, which are derived from the HL-filters, as described in 
Eq. (\ref{eq:gx3}). 
In the context of a discrete framework, this signal can be represented as a matrix, denoted by 
$\mymatrix{k}\in \mathbb{R}^{n_k\times d}$, 
where $n_k$ is the number of $k$-simplices and $d$ is the number of features. To capture the interaction and integration between $\mymatrix{\ka}$ and $\mymatrix{\kb}$~$(\ka < \kb)$, 
we utilize two fully connected layers, with a ReLU activation layer followed by the first fully connected layer, which is written as
\begin{subequations}
\begin{align}
    \mymatprm{\ka} &=
    \mathrm{ReLU}\left( ( 
        \mymatrix{\ka} \| 
            \opsstdw{\kb}{\ka}
        \mymatrix{\kb} 
        )\myweight{\ka}{\prime}
    \right) 
    \myweight{\ka}{}
    \,,\\
    \mymatprm{\kb} &=
    \mathrm{ReLU}\left( (
        \mymatrix{\kb} \| 
            \opsstup{\ka}{\kb}
        \mymatrix{\ka}
        )\myweight{\kb}{\prime}
    \right) 
    \myweight{\kb}{}
    \,,
\end{align}%
\label{tilde_xk}%
\end{subequations}%
where $(\mymatrix{a} \| \mymatrix{b})$ indicates the concatenation of $\mymatrix{a}$ and $\mymatrix{b}$, and both $\myweight{k}{\prime} \in\mathbb{R}^{2d\times d}$ and $\myweight{k}{} \in\mathbb{R}^{d\times d}$ denote weight matrices. 
Note that weight matrices are usually viewed as learnable parameters in the training.

\subsection{Simplicial Attention Pooling (SAP)} 
We introduce an innovative simplicial attention pooling technique,  encompassing  self- and cross-attention mechanisms (Fig. \ref{fig:architecture}(C)),  simplex downsampling (Fig. \ref{fig:architecture}(B)),  and attention-weighted pooling.

Our self- and cross-attention mechanisms are designed to evaluate the significance of a signal within a specific-dimensional simplex while simultaneously recognizing its topological relevance across different-dimensional simplices.  For self-attention,  we adopt a conventional transformer \cite{vaswani2017attention},  which calculates key and query weights to facilitate the learning of signal representations within specific-dimensional simplices. 

Furthermore,  by drawing upon the projection operators described in Eqs. \eqref{eq:T12} and \eqref{eq:T21},  we construct a transformer to assess the similarity between signals on $\ka$-th and $\kb$-simplices.  As a result, the overall weights assigned to the signals on the $\ka$-th and $\kb$-simplices can be expressed as a combination of their respective learned weights from both self-attention and cross-attention mechanisms.  These can be represented as:
\begin{subequations}
\begin{align}
    \mymatatt{\ka} =& 
    \mathrm{diag}\bigg(
    S\bigg(
        \alpha_{\ka}\tfrac{
            \mymatrix{\ka}\myweight{\ka}{\text{Q}} (
            \mymatrix{\ka}\myweight{\ka}{\text{K}} )^\top
        }{ \sqrt{d_k} }+ \nonumber\\&
        \hspace{4em}
        (1-\alpha_{\ka})\tfrac{
            \opsstdw{\kb}{\ka} (
            \mymatrix{\kb}\myweight{\kb}{\text{Q}} )(
            \mymatrix{\ka}\myweight{\ka}{\text{K}} )^\top
        }{ \sqrt{d_k} }
    \bigg) \bigg) \,,\label{a_k1}\\
    \mymatatt{\kb} =& 
    \mathrm{diag}\bigg(
    S\bigg(
        \alpha_{\kb}\tfrac{
            \mymatrix{\kb}\myweight{\kb}{\text{Q}} (
            \mymatrix{\kb}\myweight{\kb}{\text{K}} )^\top
        }{\sqrt{d_k}}+ \nonumber\\&
        \hspace{4em}
        (1-\alpha_{\kb})\tfrac{
            \opsstup{\ka}{\kb} (
            \mymatrix{\ka}\myweight{\ka}{\text{Q}} )(
            \mymatrix{\kb}\myweight{\kb}{\text{K}} )^\top
        }{\sqrt{d_k}}
    \bigg) \bigg) \,,\label{a_k2}
\end{align}%
\label{eq:ak}%
\end{subequations}%
where $\myweight{k}{\text{Q}}$ and $\myweight{k}{\text{K}}$ represent the query weight matrix and key weight matrix, both with dimensions of  $d\times d_k$, respectively.  Here,  $S(\cdot)$ denotes a softmax function;  $\alpha_k$ is a scalar determining the relative importance of self- and cross-attention; $ \mymatatt{\ka}$ and  $\mymatatt{\kb}$ are used as weights in the following pooling operation.

We now introduce a genetic pooling procedure for a simplicial complex denoted as $\mygraph$,  enabling simplex downsampling,  boundary operator update,  and attention-weighted pooling.  The downsampled complex is represented as $\mygraph^\prime$,  preserving simplices up to the $K$-th order.  As shown in Fig. \ref{fig:simplex}(C),  our downsampling procedure initially clusters the nodes in $\mygraph$ (i.e., $0$-simplex), which can be achieved through existing node clustering algorithms.  In this study,  we utilize the Graclus multi-level clustering algorithm \cite{dhillon2007weighted},  grouping the nodes in $\mygraph$ based on a local normalized cut.  The nodes in the same cluster are condensed into a single node in $\mygraph^\prime$ and its corresponding features are computed as the weighted average of the features defined on these nodes.  These weights are determined by $\mymatatt{0}$ as defined in Eq.  (\ref{eq:ak}).  

For further downsampling of higher-dimensional simplices, we introduce a simplex assignment matrix denoted as $\genasgn{k}$. This matrix has rows corresponding to the $k$-simplex in $\mygraph^\prime$ and columns corresponding to the $k$-simplex in $\mygraph$. Each element within $\genasgn{k}$ is binary, 
indicating whether or not the $k$-simplex in $\mygraph^\prime$ corresponds to one $k$-simplex in $\mygraph$. 
Thus, $\genasgn{k}$ characterizes the relationship of the $k$-simplex between $\mygraph$ and $\mygraph^\prime$.   


To update the boundary operator $\genopea{k}{}$, we denote the updated boundary operator as $\genopea{k}{\prime}$ and define it as follows:
\begin{equation}
    \genopea{k}{\prime}= \opbound{j}( \genasgn{k-1}\genopea{k}{}) \,,
    \label{eq:downsample}
\end{equation}
where $\opbound{j}(A)$ is an operator that removes the $j$-th column from a matrix $A$. 
 Eq.~(\ref{eq:downsample}) can be applied on any $k$-simplices by three steps: 
1) initializing the $k$-th updated boundary operator $\genopea{k}{\prime}$ to $\genasgn{k-1}\genopea{k}{}$; 
2) eliminating the $k$-simplices that contain nodes in the same node clusters; and 
3) eliminating the replicated $k$-simplices.  Fig. \ref{fig:simplex}(C) illustrates an example of the simplex downsampling up to $2-$simplex.  In this process,  while removing the $k$-simplex,  its features are aggregated into the simplex connecting to it through weighted averaging,  where the weights are defined via attention mechanisms in Eq. (\ref{eq:ak}). While $\genasgn{0}$ is computed by the node clustering algorithm, the higher-dimensional simplex assignment matrix $\genasgn{k}$ in Eq.~(\ref{eq:downsample}) can be derived from the updated boundary operator iteratively. We calculate the element of $\genasgn{k}$ based on 
$\genopea{k}{\prime}$, $\genopea{k}{}$, and $\genasgn{k-1}$, that is, 
\begin{equation}
    \begin{bmatrix}\genasgn{k}\end{bmatrix}_{ij}=
    \begin{cases}
        1 \,,&\text{if } 
        \sum_{q=1}^{n_{k-1}^\prime}
        \left| \begin{bmatrix}\genopea{k}{\prime}\end{bmatrix}_{qi} \right|
        \left| \begin{bmatrix}\genasgn{k-1}\genopea{k}{}\end{bmatrix}_{qj} \right|
        =k+1 \,;\\
        0 \,,&\text{otherwise} \,,
    \end{cases}%
    \label{eq:simassign}
\end{equation}%
where $n_{k-1}^\prime$ is the number of the $(k-1)$-simplices in the coarsened $\mygraph^\prime$. In sum, the proposed simplex downsampling algorithm can be easily implemented by first clustering nodes and computing $\genasgn{0}$. Subsequently, we update the boundary operator by Eq.~(\ref{eq:downsample}) and then compute the simplex assignment matrix with Eq.~(\ref{eq:simassign}) iteratively until $k$ reaches $K$.

\subsection{Architecture of HL-HGAT}

\begin{figure*}[htbp]
\begin{center}
\includegraphics[width=0.97\textwidth]{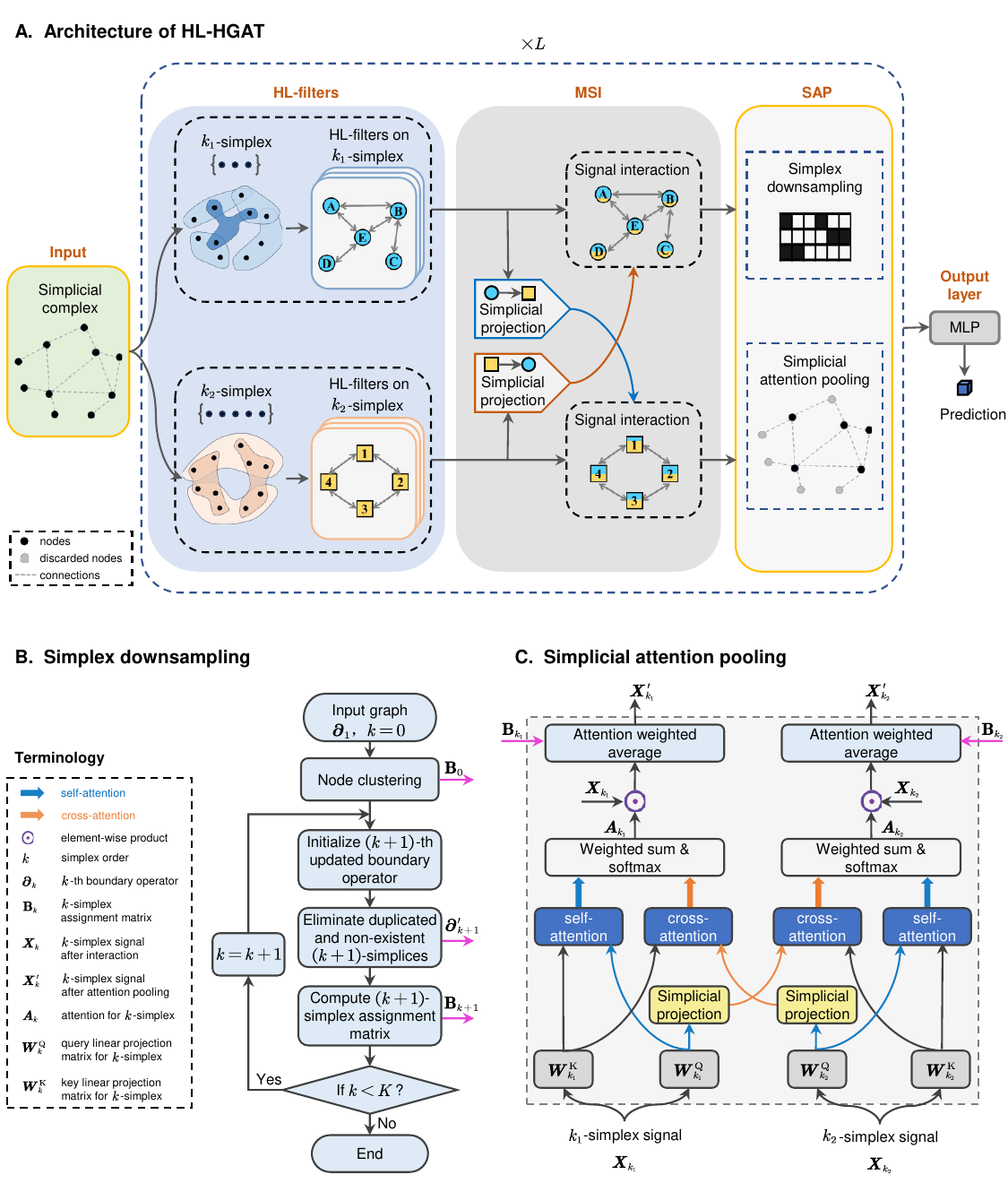}
\caption{HL-HGAT architecture.  
(A) Schematic representation of the Hodge-Laplacian Heterogeneous Graph Attention Network (HL-HGAT) architecture showcasing three key innovations: HL-filters, multi-simplicial interaction (MSI), and simplicial attention pooling (SAP). In each processing block, we initiate the workflow by applying HL-filters to signals from the $k_1$- and $k_2$-simplices from the preceding block. Subsequently, an MSI layer is employed to capture signal interactions between the $k_1$- and $k_2$-simplices. Following this, we implement an SAP layer, which involves updating the boundary operator and feature consolidation based on simplex attention. Finally, an output layer is designed for prediction.
(B) Flow chart outlining the proposed simplex downsampling algorithm in Section~\ref{appx:msi}. We employ the Graclus clustering algorithm \cite{dhillon2007weighted} to derive the node assignment matrix. This is followed by an iterative three-step process (depicted in Fig. \ref{fig:simplex}(C)): 1) Initialization of the updated boundary operator for the $(k+1)$-th iteration; 2) Removal of non-existent $(k+1)$-simplices that contain nodes within the same node clusters and duplicated $(k+1)$-simplices; 3) Computation of the $(k+1)$-simplex assignment matrix using the updated boundary operator.
(C) Schematic diagram illustrating the architecture of the Simplicial Attention Pooling (SAP). Within this framework, we compute self-attention and cross-attention for each simplex. The simplex signals are then modulated by attention mechanisms and pooled based on the assignment matrices.
} 
\label{fig:architecture} 
\end{center}
\end{figure*}

HL-HGAT is an architecture characterized by a harmonious integration of three essential components: Hodge Laplacian spectral filters (HL-filters), multi-simplicial interaction (MSI), and simplicial attention pooling (SAP), each contributing significantly to its functionality.  

Fig. \ref{fig:architecture}(A) illustrates a specific design of the HL-HGAT architecture, which can be constructed with multiple blocks, each composed of these three components. Within each block, there are convolutional branches, each equipped with HL-filters meticulously designed to process signals on dedicated dimensional simplices.  
Subsequently, the MSI module plays a pivotal role in orchestrating the interaction of signals across distinct dimensional simplices, facilitating a more comprehensive and holistic understanding of the data. 
Finally, within the architecture,  SAP serves a dual purpose: It efficiently reduces the spatial dimension of simplices while simultaneously performing information pooling, as illustrated in Fig. \ref{fig:architecture}(B) and Fig. \ref{fig:architecture}(C). 
It is worth noting that SAP may become particularly beneficial when dealing with large graph sizes. 

HL-HGAT optimally leverages the combination of these three components, working in synergy within a multi-layered architecture.  This approach is empowered to perform effective signal processing and feature extraction, making it a potent tool for a wide range of applications. 

\subsection{Implementation}
\label{sec:implementation}
As depicted in Fig. \ref{fig:architecture}, the HL-HGAT architecture comprises an output layer and multiple blocks, of which each combines HL-filter layers, MSI, and SAP. To enhance the model's effectiveness, we incorporate positional encoding for individual dimensional simplices, a practice commonly employed in existing graph transformers \cite{rampavsek2022recipe, dwivedi2020benchmarking}. Specifically, we take the Laplacian node positional encoding with sign flipping \cite{rampavsek2022recipe} as our node position encoder.  Similarly,  for edge positional encoding, we use the first eight eigenvectors of the $1$-st HL operator. This approach can be readily extended to accommodate positional encoding for simplices of any order.

HL-HGAT relies on several essential model parameters,  including: 1) the number of blocks,  2) the number of convolutional layers, 3) the number of the HL-filters per convolutional layer,  4) the polynomial order for the HL-filter approximation,  5) the size of query and key matrices in the SAP layer,  6) the number of fully connected layers in the output layer,  7) the number of points in the fully connected layers,  and 8) a dropout rate.  
For consistency across all experiments in this study, we have set the dropout rate for each convolutional layer and fully connected layer to be 0.25. 
Activation functions are chosen from either a standard ReLU or a Leaky ReLU with a leak rate of 0.1. 
Furthermore, the dimension of the query and key matrices in the SAP layer is consistently configured as $32 \times 32$. The remaining model parameters are determined through a greedy search specific to each individual application, detailed in Table~\ref{tab:parameter}. 

\begin{table*}[!t]
\centering
\caption{HL-HGAT parameters for each dataset. }
\label{tab:parameter}
\resizebox{0.99\textwidth}{!}{
\begin{tabular}{ccccccc}
    \toprule
    \textbf{Parameters} & \textbf{TSP} & \textbf{CIFAR10} & \textbf{Peptide-func} & \textbf{ZINC} & \textbf{OASIS-3} & \textbf{ABCD}\\
    \midrule
    Number of blocks & 3 & 3 & 3 & 3 & 2 & 2\\	
    Number of convolutional layers in each block & \{4, 4, 4\} & \{2, 2, 2\} & \{2, 2, 2\} & \{2, 3, 3\} & \{1, 1\} & \{2, 2\} \\
    Number of HL-filters per convolutional layer & \{32, 64, 128\} & \{64, 128, 256\} & \{64, 128, 256\} & \{64, 128, 256\} & \{32, 64\} & \{32,64\} \\
    Polynomial order & 3 & 5 & 5 & 5 & 3 & 3 \\
    Dimension of query and key per node/edge & 64 & 128 & 64 & 64 & 64 & 64 \\
    Number of FC layers in the output layer & 2 & 2 & 2 & 3 & 2 & 2 \\
    Output dimension of FC in the output layer & \{256, 1\} & \{256, 10\} & \{256, 10\} & \{256, 256, 1\} & \{256, 1\} & \{256, 1\} \\
    \bottomrule
\end{tabular}
}
\end{table*}

All experiments adhere to a consistent set of training parameters, which includes an initial learning rate of 0.001 and a weight decay parameter of 0.001. We employ the ADAM optimizer \cite{kingma2014adam} and use a mini-batch size of 64, although we may reduce the mini-batch size to 32 in cases where memory demands exceed GPU capacity.

HL-HGAT is implemented using Python version 3.9.13 and relies on the PyTorch 1.12.1 framework \cite{paszke2019pytorch} along with the PyTorch Geometric 2.1.0 library \cite{pyg}. The execution of HL-HGAT is carried out on an NVIDIA Tesla V100SXM2 GPU boasting 32GB of RAM.


\section{Experiments}
\label{sec:empirical}

In the following, we demonstrate the use of the proposed HL-HGAT in NP-hard combinatorial optimization problems, graph classification, and multi-label tasks, as well as graph regression tasks. To illustrate the versatility of HL-HGAT, we conduct experiments using established open datasets across various domains,  including logistics,  computer vision,  biology, chemistry,  and neuroscience.  These demonstrations showcase the adaptability and effectiveness of HL-HGAT in diverse real-world applications.

\subsection{Traveling Salesman Problem}  

The Traveling Salesman Problem (TSP) represents a classic NP-hard challenge within combinatorial optimization.  It revolves around determining the most efficient route to visit all cities exactly once on a map, given a list of cities and the distances between each pair of them. The importance of solving the TSP lies in its wide array of practical applications, such as route planning for logistics and delivery services \cite{gavish1978travelling}. The TSP also serves as a fundamental model for more complex optimization problems, making advancements in its solution methods applicable to a broader range of challenges \cite{johnson1997traveling}. 

In this study,  we model each map as a graph, with cities represented as nodes and the connections between them as edges.  A total of 12,000 maps\footnote{\url{https://www.math.uwaterloo.ca/tsp/index.html}} were employed in this research, each featuring a varying number of nodes, ranging from 50 to 500 nodes.  Each node is connected to 25 other nodes through edges.
The HL-HGAT approach tackles the TSP by framing it as an edge classification task.  This is achieved through the minimization of a focal loss \cite{lin2017focal},  which effectively quantifies the weighted entropy of edge probabilities.  To estimate the probability of each edge within the shortest path, the HL-HGAT architecture omits the SAP component.  Input features encompass the coordinates of cities assigned to nodes and the distances between all pairs of cities assigned to edges. We meticulously follow the dataset-splitting strategy outlined in the benchmarking paper by Dwivedi \emph{et al.} \cite{dwivedi2020benchmarking}, which includes 10,000 training graphs, 1,000 validation graphs, and 1,000 test graphs.  The results presented below were obtained from the test set.

\begin{figure*}[htbp]
\begin{center}
\includegraphics[width=0.94\textwidth]{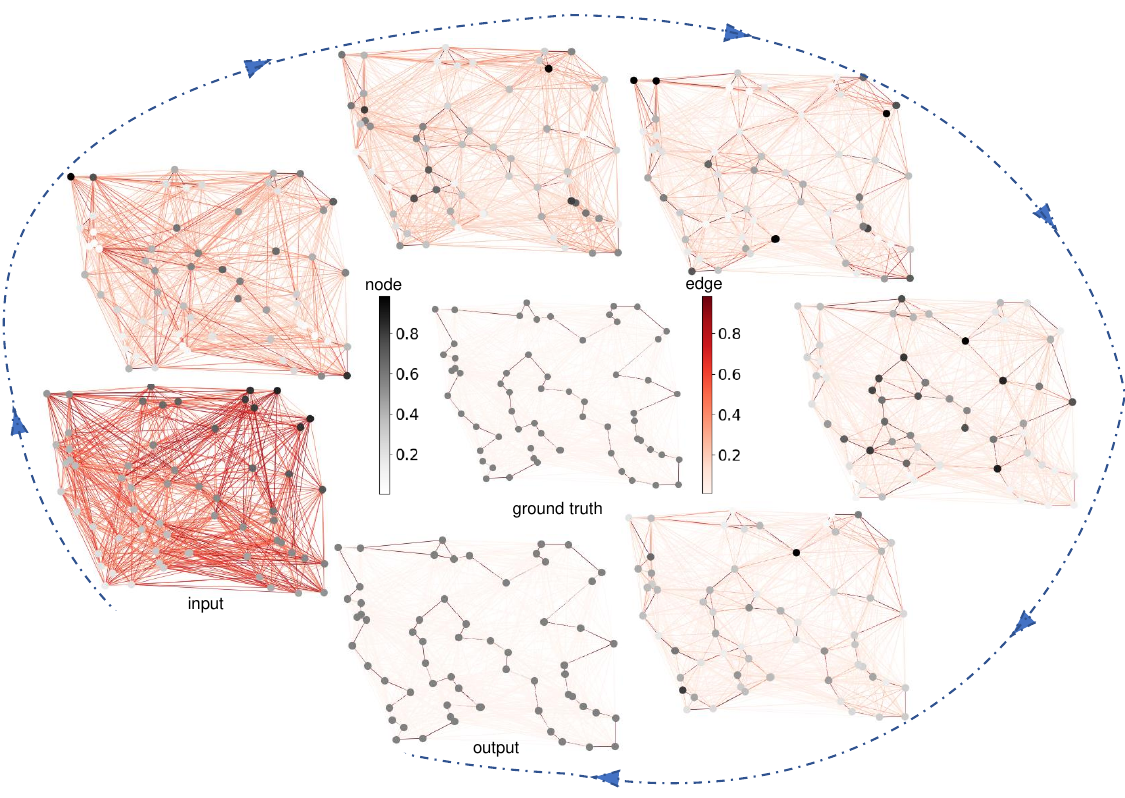}
\caption{Traveling Salesman Problem (TSP). Node and edge features at every two HL-filter layers are visualized. The ground truth is positioned at the center of the figure. Node features are represented in grey, while edge features are highlighted in red.}  
\label{fig:tsp_trend} 
\end{center}
\end{figure*}

\begin{figure*}[!htbp]
\begin{center}
\includegraphics[width=0.94\textwidth]{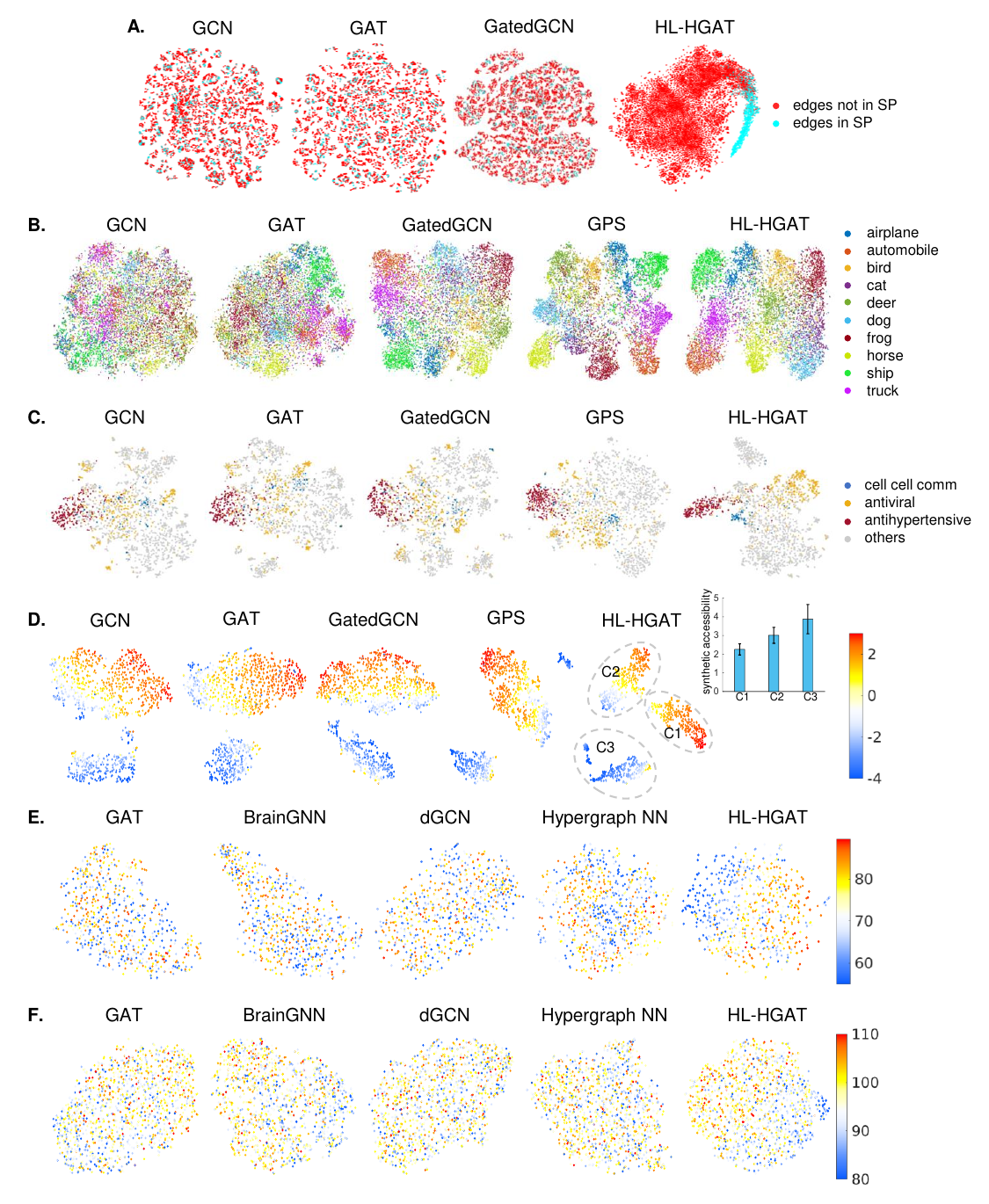}
\caption{Feature space.  The feature space learned from each GNN method is projected into a two-dimensional space using t-distributed stochastic neighbor embedding (t-SNE) for visualization purposes.  
(A) In the feature space of the Traveling Salesman Problem,  red points indicate edges that do not belong to the shortest path,  while cyan points represent edges that are part of the shortest path. 
(B) The feature space of the CIFAR10 dataset visualizes the 10 classes of natural images. 
(C) In the feature space of the Peptide-func dataset,  the colored points indicate peptides with corresponding functions, while grey points indicate peptides that do not have these functions. 
(D) In the feature space of the ZINC dataset,  the points represent molecules colored by the corresponding constrained solubility.  Additionally,  the bar plot illustrates the average and standard deviation of synthetic accessibility across all molecules in the three clusters.
(E--F) The feature spaces of brain images are color-coded by brain age and general intelligence.  Each point in these plots represents an individual subject.  Each column corresponds to the feature space obtained from various GNN models, including GCN \cite{kipf2016semi}, GAT \cite{velivckovic2017graph},  GatedGCN \cite{bresson2017residual},  GPS\cite{rampavsek2022recipe}, dGCN \cite{zhao2022dynamic}, BrainGNN \cite{li2021braingnn}, and Hypergraph NN \cite{jo2021edge}.  
}  
\label{fig:tsne} 
\end{center}
\end{figure*}

Fig. \ref{fig:tsp_trend} showcases the visualization of node and edge features every two layers, illustrating how the proposed HL-filters effectively eliminate irrelevant edge signals.  In the supplementary material, we present a range of TSP examples featuring varying numbers of cities and their optimized solutions generated by HL-HGAT, along with comparisons to several state-of-the-art node-centric GNN models, including GCN \cite{kipf2016semi}, GAT \cite{velivckovic2017graph}, and GatedGCN \cite{bresson2017residual}. Visually,  our HL-HGAT in the supplementary material offers solutions that closely align with the ground truth, outperforming GCN, GAT, and GatedGCN. 
This phenomenon can also be observed in the feature space learned from each method (i.e., Fig. \ref{fig:tsne}(A)),  suggesting a clear separation of edges belonging to the shortest path and edges not belonging to the path.

\begin{figure*}[!htbp]
\begin{center}
\includegraphics[width=0.92\textwidth]{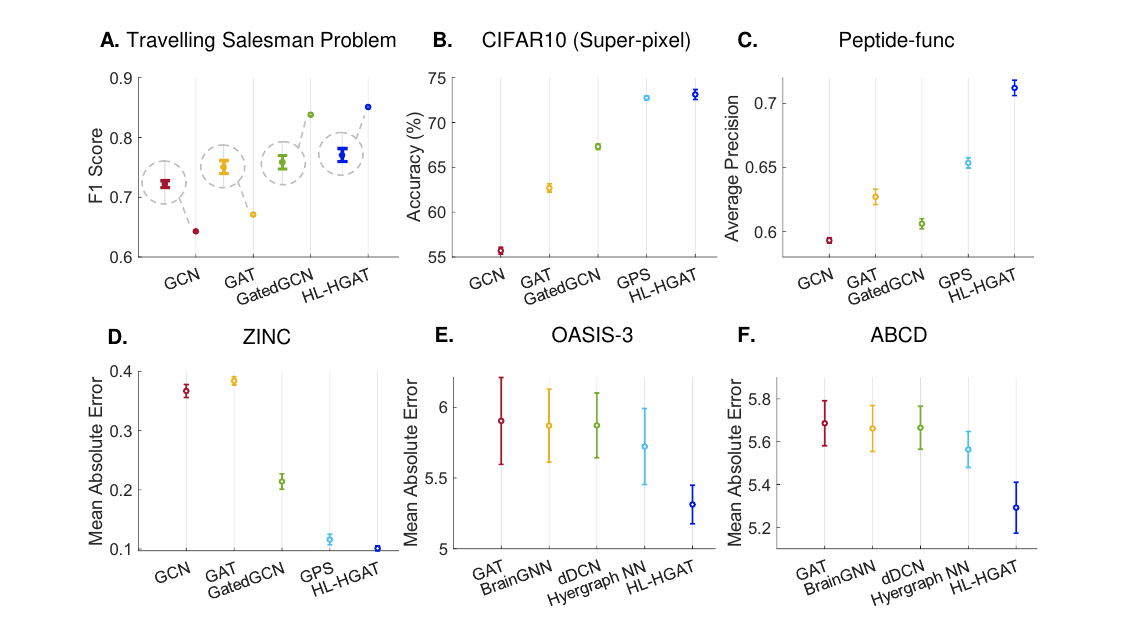}
\caption{Quantitative comparisons between HL-HGAT and state-of-the-art GNN methods across multiple datasets.  Error bars illustrating quantitative metrics are presented in panels (A--F) for various datasets, including TSP, CIFAR10, Peptide-func, ZINC, OASIS-3, and ABCD. Notably,  panel (A) zooms in on enlarged error bars enclosed within gray dashed circles. These quantitative metrics have been selected based on commonly reported measures in the existing literature \cite{dwivedi2020benchmarking, dwivedi2022long}.}
\label{fig:comparison_errbar} 
\end{center}
\end{figure*}

Quantitatively, Fig. \ref{fig:comparison_errbar}(A) illustrates the edge classification accuracy evaluated by F1 score. Two-sample $t$-tests show that the F1 score from HL-HGAT is significantly greater than GCN, GAT, and GatedGCN ($p<0.001$),  suggesting the superior performance of the HL-HGAT over the state-of-art GNN methods.

\subsection{Image Classification} 
Image classification is a typical task in computer vision, with applications in numerous areas including medical imaging, surveillance, autonomous vehicles, etc \cite{lu2007survey}.  The HL-HGAT approach treats classical natural image classification problems as graph multi-class classification tasks. 

We employ the CIFAR10 dataset of RGB images 
(with 60,000 graphs and 10 distinct classes), 
which is a well-known dataset in the field of computer vision \cite{krizhevsky2009learning}. To facilitate graph-based representation,  each superpixel, defined as a compact and contiguous region in images showcasing uniform intensity characteristics,  is treated as a node \cite{achanta2012slic}, and they are connected to their eight nearest neighbors based on the Euclidean distance. On average,  each graph in the dataset consists of 117 nodes. This forms a graph where superpixels are nodes, and edges connect neighboring superpixels. The nodes incorporate RGB values, and edges are established based on the absolute differences and Euclidean distances between corresponding RGB color values. These signal attributes enable the model to effectively capture both color information and spatial relationships among the superpixels within an image.  In this experiment,  we follow the dataset splitting strategy as described in Dwivedi et al. \cite{dwivedi2020benchmarking}, which involves dividing the dataset into 45,000 training graphs, 5,000 validation graphs, and 10,000 test graphs. Notably,  all three key components are integrated into the architecture of HL-HGAT for this specific application.

Fig. \ref{fig:tsne}(B) illustrates the two-dimensional embedding of the features obtained through GCN \cite{kipf2016semi}, GAT \cite{velivckovic2017graph},  GatedGCN \cite{bresson2017residual},  GPS \cite{rampavsek2022recipe},  and our HL-HGAT.  Each point in this figure corresponds to an individual image, with color coding indicating its respective class. Notably, HL-HGAT achieves a more distinct separation of the 10 classes when compared to the other methods.  

Quantitatively, Fig. \ref{fig:comparison_errbar}(B) presents classification accuracy, which measures the proportion of correctly classified images relative to the total dataset size.  
The results of two-sample $t$-tests reveal that the accuracy achieved by HL-HGAT significantly surpasses those of other GNN models (such as GCN, GAT, and GatedGCN) as well as the graph transformer model (GPS) (all $p$-values $<0.01$). These results suggest the effectiveness of HL-HGAT in computer vision tasks, establishing its superiority over the state-of-the-art GNN and graph transformer models.

\subsection{Biology and Chemistry} 
The study employs two widely-used molecular datasets,  namely ZINC \cite{irwin2012zinc} and Peptide-func \cite{singh2016satpdb}, to demonstrate the use of HL-HGAT in the fields of biology and chemistry. 

The Peptide-func dataset 
with 15,535 graphs 
is specifically tailored for predicting peptide functions that are crucial for drug discovery, biological signaling, functional genomic discovery, etc \cite{wieder2020compact}.  The study considers this task as a multi-label graph classification problem that assigns each peptide with multiple functions, such as antibacterial, antiviral, intercellular communication, and more. Peptides, which are short chains of amino acids, are derived from SATPdb, a reference database \cite{singh2016satpdb}. In contrast to conventional approaches where amino acids typically serve as nodes in peptide graphs \cite{borgwardt2005protein},  this study employs heavy atoms as nodes, effectively expanding the size of the graph. On average, each graph in the dataset consists of 151 nodes. The edges within these graphs represent the bonds between these heavy atoms.  

Similarly,  ZINC is a curated repository of chemical compounds for virtual screenings of drug discovery, structural bioinformatics, and chemoinformatics research.  We focus on a carefully selected subset of ZINC molecular graphs, consisting of 12,000 samples \cite{irwin2012zinc}. Our primary objective is to perform regression analysis on a molecular property referred to as ``constrained solubility'' that is crucial in determining drug efficacy and safety in pharmaceutical development \cite{cumming2017octanol}. Accurate solubility prediction enables the design of drugs with optimal absorption and bioavailability, ensuring effective treatment and minimizing potential side effects \cite{icsik2020octanol}. This task constitutes a typical graph regression problem, with our evaluation metric being mean absolute error (MAE).  In each of these molecular graphs, the node features represent the types of heavy atoms present, while the edge features characterize the types of bonds connecting these atoms. It is worth noting that the graphs in our dataset exhibit varying sizes, with node counts ranging from 9 to 37 and edge counts ranging from 16 to 84.  We adhere to the dataset splitting strategy outlined in Dwivedi et al. \cite{dwivedi2022long}, which involves a train-validation-test split ratio of 70\%--15\%--15\%. 

Fig. \ref{fig:tsne}(C) provides a visual representation of the two-dimensional feature embeddings obtained through GCN \cite{kipf2016semi}, GAT \cite{velivckovic2017graph}, GatedGCN \cite{bresson2017residual}, GPS \cite{rampavsek2022recipe}, and our HL-HGAT. In this plot, each point represents an individual peptide, with color coding indicating its corresponding peptide function. 

Similarly,  Fig. \ref{fig:tsne}(D) provides a visual representation of the two-dimensional feature embeddings of the ZINC dataset obtained through GCN \cite{kipf2016semi}, GAT \cite{velivckovic2017graph}, GatedGCN \cite{bresson2017residual}, GPS \cite{rampavsek2022recipe}, and our HL-HGAT. Each data point in this plot corresponds to an individual chemical compound, and its color code indicates its constrained solubility.  Significantly, the embedding plot generated by HL-HGAT reveals three distinct clusters among these chemical compounds, whereas most of the other compared GNN and transformer models exhibit only two clusters.  Further exploration of the three clusters identified by HL-HGAT reveals a connection to the synthetic accessibility score (SAS). The SAS evaluates the structural characteristics of drug-like molecules \cite{ertl2009estimation}, as illustrated in the bar chart within the last panel of Fig. \ref{fig:tsne}(D). This valuable insight into the SAS-related clustering dimension is a unique discovery attributed to HL-HGAT beyond the objective of this prediction task and is not observed in the other GNN models and GPS.

Quantitatively,  the results of two-sample $t$-tests reveal that the average precision achieved for the Peptide-func dataset and the mean absolute error (MAE) obtained for the ZINC dataset through HL-HGAT demonstrate significantly superior classification or prediction accuracy when compared to GCN \cite{kipf2016semi}, GAT \cite{velivckovic2017graph}, GatedGCN \cite{bresson2017residual}, and GPS \cite{rampavsek2022recipe} (all $p$-values $<0.01$).

\begin{figure*}[htbp]
\begin{center}
\includegraphics[width=0.94\textwidth]{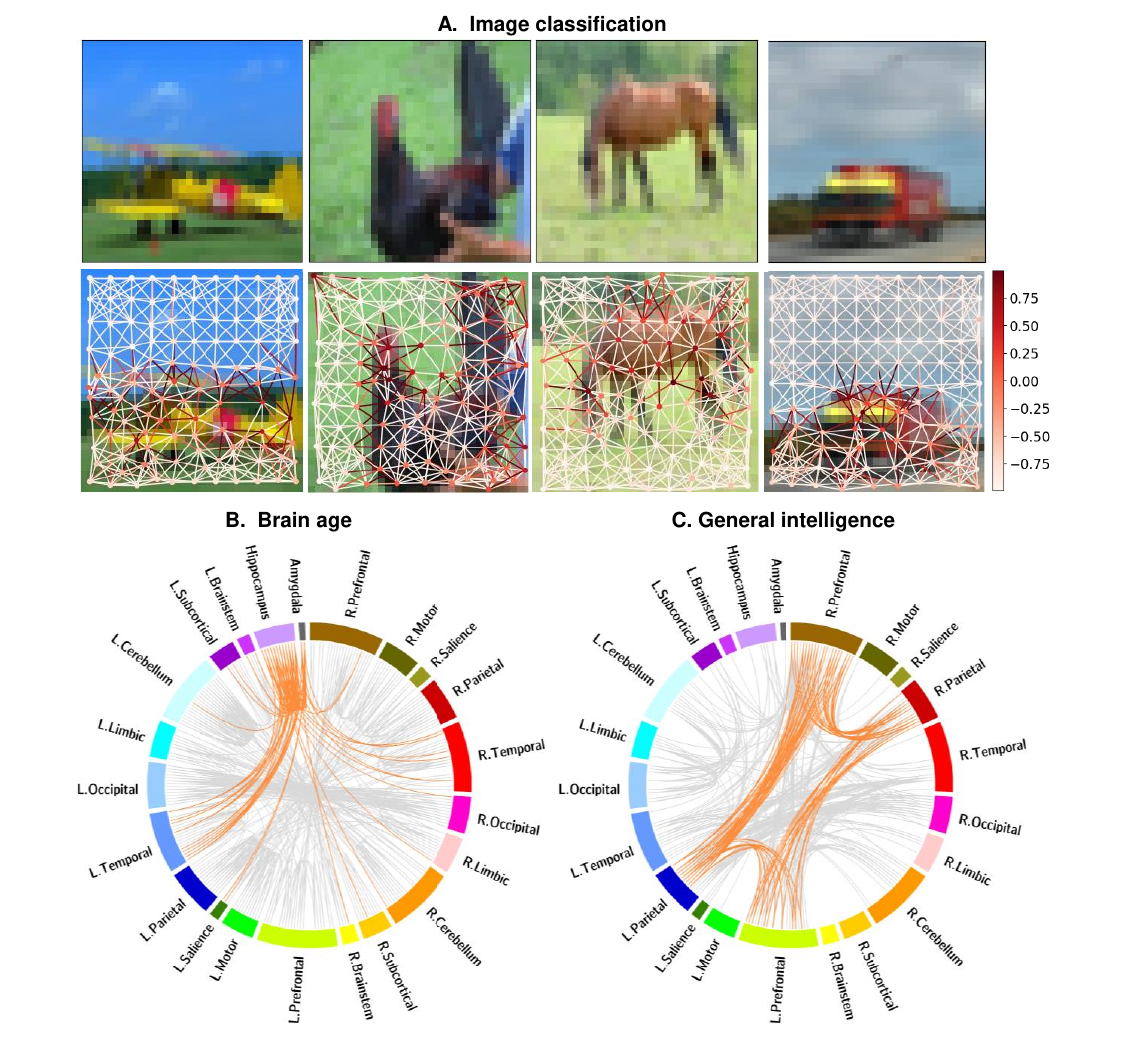}
\caption{ Attention maps obtained from the SAP component in HL-HGAT for image classification. (A) Image Classification: Attention maps highlight nodes and edges with increased attention in red, emphasizing key features and boundaries crucial for image classification. (B) Brain Age Prediction: The attention maps visualize the functional connectivities (colored in orange) most contributing to age prediction. (C) General Intelligence prediction: Similarly, the functional connectivities colored in orange are most contributed to the prediction of general intelligence. Abbreviations: L, left; R, right.}  
\label{fig:attmap} 
\end{center}
\end{figure*} 

\subsection{Brain Age and Intelligence} 

Two brain functional MRI datasets, derived from the Adolescent Brain Cognitive Development Study (ABCD)\footnote{\url{https://abcdstudy.org/}} \cite{BARCH201855} and Open Access Series of Imaging Studies (OASIS-3)\footnote{\url{https://www.oasis-brains.org}} \cite{marcus2010open},  
are used in this study to predict intelligence \cite{akshoomoff2013viii} and brain age,  respectively.  The sample sizes of the ABCD and OASIS-3 datasets are $7,693$ and $1,978$, respectively.  Brain age has been demonstrated to be a biomarker related to neurodegenerative diseases \cite{franke2019ten}, while intelligence is a good predictor for academic outcomes in children \cite{deary2007intelligence}.

For the purpose of constructing our brain functional organization graphs, we employ a total of 268 regions of interest (ROIs), as previously defined by Shen et al. (2017)  \cite{shen2017using}. These ROIs serve as the nodes within our graph, and we establish edges between them based on Pearson correlation calculated from the functional time series of each pair of ROIs.  The sparsity level of our graph is set at 0.25.  We employ the brain functional networks to predict brain age and intelligence. Our experimental setup closely follows the methodology outlined in Huang et al. \cite{huang2023heterogeneous}.

Figures \ref{fig:tsne}(E) and \ref{fig:tsne}(F) 
depict the two-dimensional feature embeddings obtained from GAT \cite{velivckovic2017graph}, BrainGNN \cite{li2021braingnn},  dGCN \cite{zhao2022dynamic}, Hypergraph NN \cite{jo2021edge}, and our HL-HGAT.  Notably, the selection of comparison graph models in this context differs from those mentioned earlier. BrainGNN \cite{li2021braingnn},  dGCN \cite{zhao2022dynamic}, and Hypergraph NN \cite{jo2021edge} are specialized techniques tailored for brain functional networks, representing state-of-the-art approaches in the field of brain research. 
In these two sub-figures, each data point corresponds to an individual brain functional network, with color coding denoting their respective brain age and intelligence attributes.

The results, as depicted in 
Figures \ref{fig:comparison_errbar}(E) and \ref{fig:comparison_errbar}(F), 
highlight the remarkable performance achieved by our HL-HGAT, positioning it at the forefront as one of the state-of-the-art solutions on both datasets when compared to the baseline models.  For the OASIS-3 dataset, our HL-HGCNN attains state-of-the-art results in comparison to the baseline GNN models for node signals, such as GAT, BrainGNN,  and dGCN (all $p$-values $< 0.01$).  
Additionally, our model surpasses the GNN model for edge signals, as evidenced by its superior performance relative to Hypergraph NN  ($p=0.02$).  Similarly,  in the case of the ABCD dataset, rigorous two-sample $t$-tests also reveal that the mean absolute errors (MAE) obtained through HL-HGAT significantly outshine those of GAT,  BrainGNN,  dGCN,  and Hypergraph NN (all $p$-values $< 0.01$).


\begin{figure*}[!htbp]
\begin{center}
\includegraphics[width=0.92\textwidth]{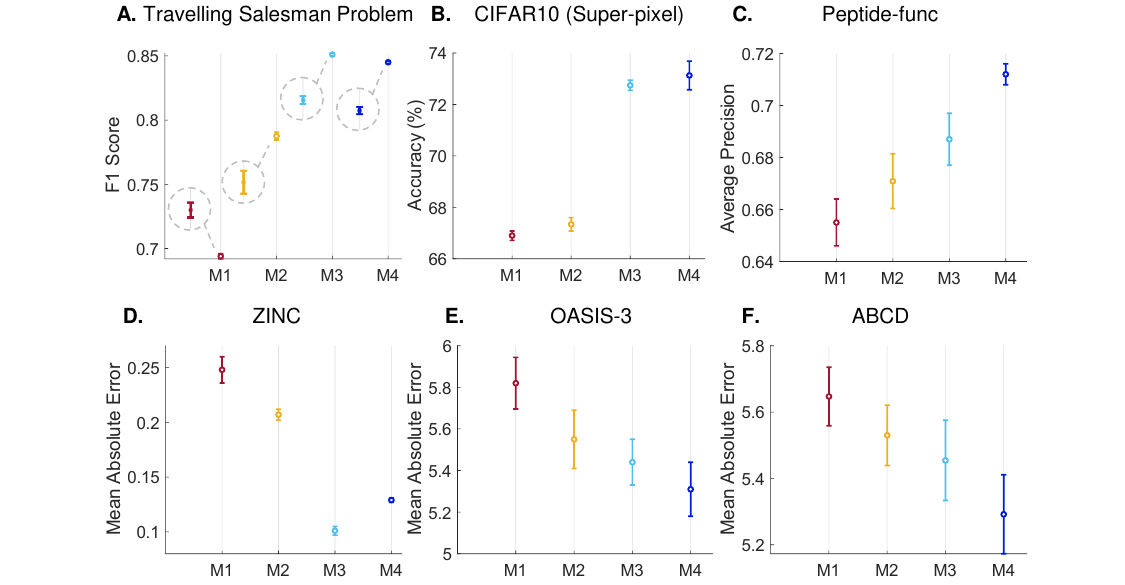}
\caption{Ablation Study.  
Panels (A--F) illustrate error bars for quantitative metrics across various datasets, including TSP, CIFAR10, Peptide-func, ZINC, OASIS-3, and ABCD. Panel (A) provides a close-up view of enlarged error bars enclosed within gray dashed circles.
Abbreviations:
M1 - HL-filters applied to node signals only;
M2 - HL-filters applied to both node and edge signals;
M3 - HL-filters applied to node and edge signals with MSI integration;
M4 - The complete architecture encompassing all three key components as depicted in Fig. \ref{fig:architecture}(A).
} 
\label{fig:ablation_errbar} 
\end{center}
\end{figure*}

\subsection{SAP Provides Interpretable Attention Maps}
The attention maps generated by the SAP component offer valuable interpretability to HL-HGAT, which is often considered opaque in existing GNN models.  In this section, we demonstrate the interpretability of HL-HGAT through attention maps for the CIFAR10,  ABCD, and OASIS datasets, originating from the SAP pooling layer.

For the image classification task, we anticipate that node attention will emphasize objects, while edge attention should accentuate object boundaries. To validate this notion, we randomly select four samples from the test set and overlay the attention onto superpixel images.  Fig. \ref{fig:attmap}(A) illustrates that SAP effectively identifies and highlights both objects and their boundaries across samples spanning various classes. Additional examples of these attention maps are available in the supplementary materials.

In the context of brain age and general intelligence prediction, the attention maps derived from brain functional networks serve as informative biomarkers, shedding light on neural mechanisms linked to aging and intelligence.  
Figures~\ref{fig:attmap}(B) and \ref{fig:attmap}(C) 
depict functional connectivities with the top 5\% attention scores. Heightened attention is notably observed in the functional connectivities of the hippocampus and amygdala for brain age prediction, as these regions play pivotal roles in memory and emotional responses, both closely tied to aging \cite{nashiro2012age, bettio2017effects}. Furthermore, our attention maps unveil significant functional connectivities between prefrontal and parietal regions that contribute to the prediction of general intelligence, aligning with established research on neural activities in these regions \cite{jung2007parieto, song2008brain}.

These findings underscore the interpretability of HL-HGAT, as it consistently provides meaningful attention maps across diverse applications, enriching our comprehension of complex data and underlying phenomena.


\subsection{Ablation Study}

To comprehensively evaluate the significance of each component within HL-HGAT, we conduct an ablation study. We systematically increase the complexity of HL-HGAT by considering different architectural variations, each represented by a model denoted as follows: M1 (HL-filters applied to node signals only), M2 (HL-filters applied to both node and edge signals), M3 (HL-filters applied to node and edge signals with MSI integration), and M4 (the full architecture encompassing all three key components in Fig. \ref{fig:architecture}(A)). It is worth noting that we maintain consistent hyperparameters, including the number of blocks, the number of HL-filters within each block, the order of Laguerre polynomials, and other learning parameters, across all experiments, as detailed in Section \ref{sec:implementation}.

Fig.  \ref{fig:ablation_errbar} clearly demonstrates that increasing the complexity of HL-HGAT from M1 to M4 consistently results in improved classification accuracy for the CIFAR10 and Peptide-func datasets 
shown in Figures \ref{fig:ablation_errbar}(B) and \ref{fig:ablation_errbar}(C), 
as well as enhanced prediction performance for brain age (i.e., OASIS-3, shown in Fig.  \ref{fig:ablation_errbar}(E)) and general intelligence (i.e., ABCD,  shown in Fig.  \ref{fig:ablation_errbar}(F)). These findings underline the critical importance of HL-filters for handling heterogeneous signals, as well as the integration of MSI and SAP for effective multi-scale information processing. However, it is noteworthy that M4 does not consistently outperform M3 in the TSP and ZINC datasets. This discrepancy can be attributed to the significantly smaller graph size in the ZINC dataset, where the graph pooling operation may not necessarily enhance multi-scale processing. Furthermore, for the TSP problem, the nature of individual edge prediction may render pooling to be unnecessary.

\begin{figure*}[htbp]
\begin{center}
\includegraphics[width=0.9\textwidth]{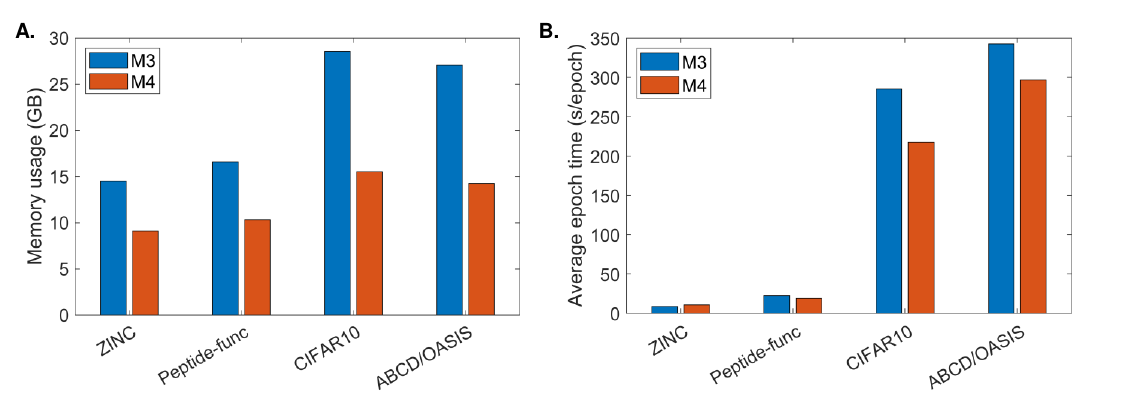}
\caption{ Impact of the SAP component on memory usage (A) and computational time (B).  The average edge counts in the ZINC, Peptide-func, CIFAR10,  and ABCD or OASIS-3 are respectively 50,  307, 941, and 8978.  Abbreviations: M3, HL-filters applied to node and edge signals with MSI integration; M4,  the full architecture encompassing all three key components in Fig. \ref{fig:architecture}(A).}  
\label{fig:pooling_effect} 
\end{center}
\end{figure*}

Nonetheless, the SAP pooling operation stands out as a valuable addition, not only improving computational efficiency but also reducing memory usage. The results presented in Fig.  \ref{fig:pooling_effect} clearly indicate that our graph pooling operator significantly reduces both memory cost and computational time, with particular advantages evident for larger graphs. This efficiency enhancement is crucial for scaling HL-HGAT to handle larger and more intricate datasets.

\section{Conclusion}
\label{sec:concl}

In this study, we introduce HL-HGAT, a versatile graph neural network designed to effectively model signals on nodes, edges, and higher-dimensional simplices within graphs. HL-HGAT harnesses the combined capabilities of HL-filters, MSI, and SAP to address complex problems involving graph-structured data. Our framework offers flexibility in combining these modules, providing a user-friendly workflow for a wide range of applications.

Our experiments, conducted on six diverse datasets spanning various domains---including NP-hard problems, graph classification, multi-label problems, and graph regression---consistently demonstrate HL-HGAT's superiority over state-of-the-art GNN and graph transformer models. This consistent performance highlights HL-HGAT as an advanced GNN model that excels in various scenarios. Furthermore, the ablation study underscores the significance of each component within HL-HGAT, emphasizing the importance of modeling heterogeneous signals, facilitating interactions across different-dimensional simplices, and leveraging the SAP layer for effective information aggregation. Additionally, the incorporation of the graph pooling operator significantly enhances the model's computational efficiency, making it well-suited for larger graph datasets.

The interpretability provided by the attention maps generated by the SAP layer further enhances the model's utility. For instance, in the CIFAR10 dataset, the attention maps successfully capture objects and their boundaries, while in brain datasets, they highlight task-related functional connectivity patterns. This interpretability holds promise for applications in medical imaging and computer vision.

However, our approach has a limitation in the form of the simplex downsampling method, which simplifies graphs solely based on their topology and may inadvertently remove significant simplices contributing to downstream tasks. In future work, we aim to develop a more robust coarsening method that considers simplicial attention to address this limitation.

In conclusion, HL-HGAT represents a powerful and flexible GNN model that is capable of solving diverse graph-based problems by effectively incorporating heterogeneous information, modeling complex interactions, and providing computational efficiency. Its interpretability further enhances its applicability across various fields, promising valuable contributions to graph-based machine learning applications.





\section*{Acknowledgments}
This research/project is supported by STI 2030—Major Projects (No.2022ZD0209000)] and the National Research Foundation, Singapore, and the Agency for Science Technology and Research (A*STAR), Singapore, under its Prenatal/Early Childhood Grant (Grant No. H22P0M0007). Additional support is provided by the Hong Kong global STEM scholar scheme and the internal fund of the Hong Kong Polytechnic University.

Data used in the preparation of this article were obtained from the Adolescent Brain Cognitive Development (ABCD) Study (https://abcdstudy.org), held in the NIMH Data Archive (NDA). This is a multisite, longitudinal study designed to recruit more than 10,000 children age 9--10 and follow them over 10 years into early adulthood. The ABCD Study is supported by the National Institutes of Health and additional federal partners under award numbers U01DA041022, U01DA041028, U01DA041048, U01DA041089, U01DA041106, U01DA041117, U01DA041120, U01DA041134, U01DA041148, U01DA041156, U01DA041174, U24DA041123, and U24DA041147. A full list of supporters is available at https://abcdstudy.org/nih-collaborators. A listing of participating sites and a complete listing of the study investigators can be found at https://abcdstudy.org/principal-investigators.html. ABCD consortium investigators designed and implemented the study and/or provided data but did not necessarily participate in analysis or writing of this report. This manuscript reflects the views of the authors and may not reflect the opinions or views of the NIH or ABCD consortium investigators.



\bibliographystyle{ieeetr}
\bibliography{Ref.bib}

\section{Biography Section}
 
\vspace{11pt}
\begin{IEEEbiography}[{\includegraphics[width=1in,height=1.25in,clip,keepaspectratio]{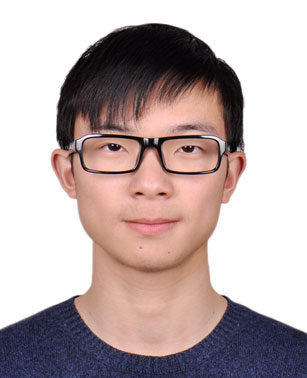}}]{Jinghan Huang}
received his BS degree from the School of Physics and Astronomy at Shanghai Jiao Tong University, China. He is currently working towards an MEng degree in Biomedical Engineering at the National University of Singapore. His current research interests are graph deep learning and its applications to medical image analysis.
\end{IEEEbiography}

\vspace{11pt}
\begin{IEEEbiography}[{\includegraphics[width=1in,height=1.25in,clip,keepaspectratio]{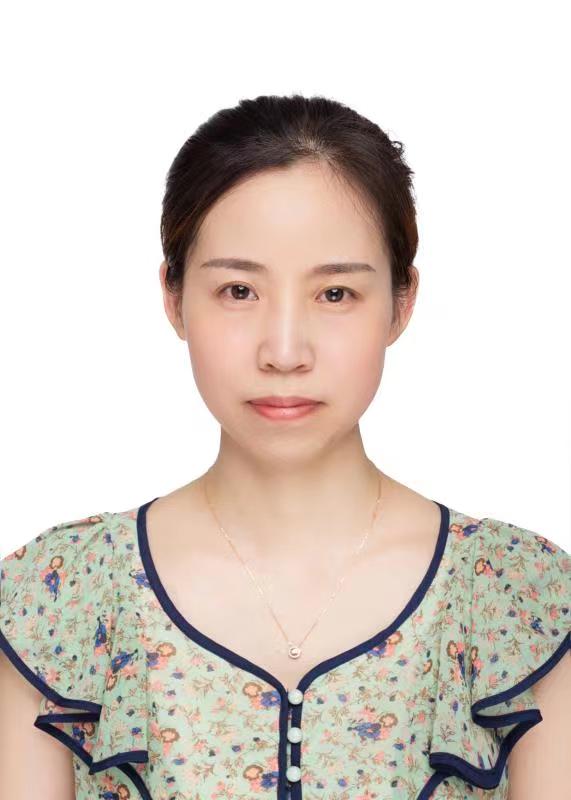}}]{Qiufeng Chen}
received her BS, MS, and PhD degrees from the School of Information Science and Engineering at Central South University, Changsha, China, in 2006, 2009, and 2016, respectively. She was a visiting scholar at the School of Biomedical Engineering, Shanghai Tech University, Shanghai, China, in 2022, and at the Department of Biomedical Engineering, National University of Singapore, Singapore, in 2023. She is currently a lecturer at the College of Computer and Information Science in Fuzhou, China. Her research mainly focuses on machine learning in medical image analysis and computer-aided diagnosis.
\end{IEEEbiography}

\vspace{11pt}
\begin{IEEEbiography}[{\includegraphics[width=1in,height=1.25in,clip,keepaspectratio]{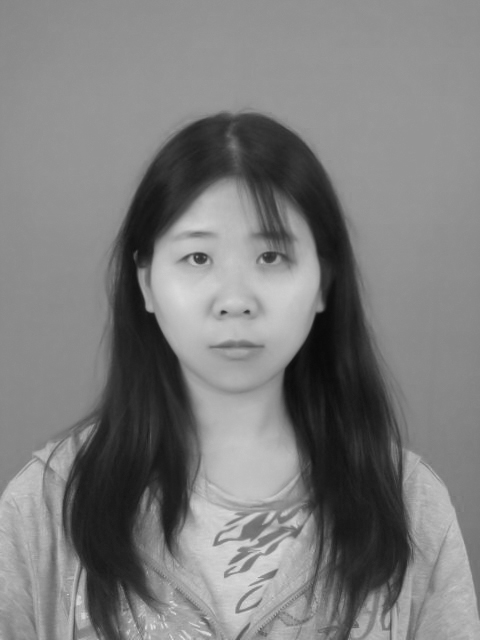}}]{Yijun Bian}
received the PhD degree in computer science and technology from the University of Science and Technology of China, Hefei, China, in 2020. She was a visiting research student with the Texas A\&M University once and is currently a research fellow with the National University of Singapore, Singapore. Her research interests include machine learning, ensemble methods, and fairness in machine learning. 
\end{IEEEbiography}

\vspace{11pt}
\begin{IEEEbiography}[{\includegraphics[width=1in,height=1.25in,clip,keepaspectratio]{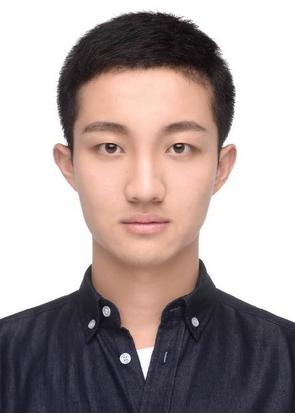}}]{Pengli Zhu}
received the B.Eng. degree in marine engineering and the M.Eng. degree in naval architecture and ocean engineering, in 2017 and 2020, respectively, from Dalian Maritime University, Dalian, China, where he is currently working toward the Ph.D. degree in marine engineering.
He is also a joint-training Ph.D. student with the College of Design and Engineering, National University of Singapore, Singapore. His current research interests include image analysis, intelligent control, and machine learning.
\end{IEEEbiography}

\begin{IEEEbiography}[{\includegraphics[width=1in,height=1.25in,clip,keepaspectratio]{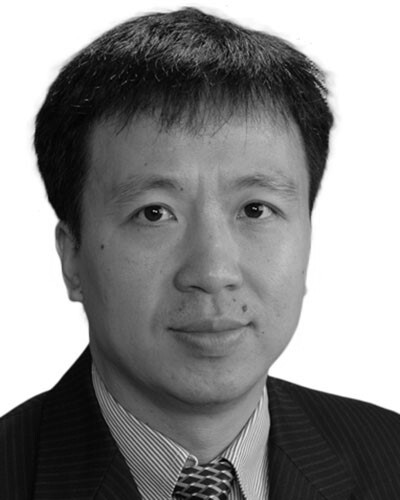}}]{Nanguang Chen} 
is an Associate Professor of Biomedical Engineering at the National University of Singapore (NUS). He received his Ph.D. in Biomedical Engineering from Tsinghua University. He also received his MSc in Physics (Peking University) and BSc in Electrical Engineering (Hunan University), respectively.  His research interests include diffuse optical tomography, optical coherence tomography, and novel fluorescence microscopic imaging methods. He has published more than 180 papers and holds five international patents.
\end{IEEEbiography}

\begin{IEEEbiography}[{\includegraphics[width=1in,height=1.25in,clip,keepaspectratio]{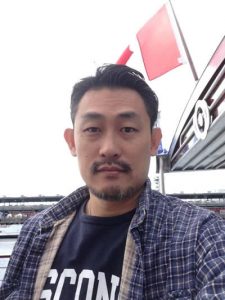}}]{Moo K. Chung} 
Ph.D., is a Professor of Biostatistics and Medical Informatics at the University of Wisconsin-Madison. He is also affiliated with the Waisman Laboratory for Brain Imaging and Behavior. Dr. Chung received Ph.D. from the Department of Mathematics at McGill University under Keith J. Worsley and James O. Ramsay. Dr. Chung’s main research area is computational neuroimaging, where noninvasive brain imaging modalities such as magnetic resonance imaging (MRI) and diffusion tensor imaging (DTI) are used to map the spatiotemporal dynamics of the human brain. His research concentrates on the methodological development required for quantifying and contrasting anatomical shape variations in both normal and clinical populations at the macroscopic level using various mathematical, statistical and computational techniques.

\end{IEEEbiography}

\begin{IEEEbiography}[{\includegraphics[width=1in,height=1.25in,clip,keepaspectratio]{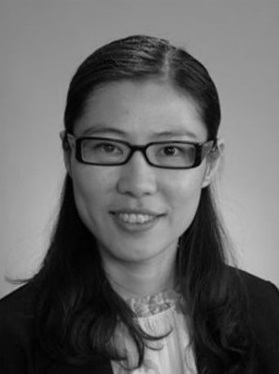}}]{Anqi Qiu} received the PhD degree in electrical and computer engineering from Johns Hopkins University. She is currently a Professor and Global STEM scholar at the Department of Health Technology and Informatics, the Hong Kong Polytechnic University. 
Dr. Qiu has been devoted to innovation in computational analyses of complex and informative datasets comprising of disease phenotypes, neuroimage, and genetic data to understand the origins of individual differences in health throughout the lifespan. 
Methodologically she focus on the spectral geometry and deep learning. 
\end{IEEEbiography}




\end{document}